\def\year{2019}\relax
\documentclass[letterpaper]{article} 
\usepackage{aaai19}  
\usepackage{times}  
\usepackage{helvet}  
\usepackage{courier}  
\usepackage{url}  
\usepackage{graphicx}  
\usepackage{amsfonts}
\usepackage{amsmath}
\usepackage{xcolor}
\graphicspath{ {images/} }
\usepackage{subcaption}
\nocopyright

\newcommand*\samethanks[1][\value{footnote}]{\footnotemark[#1]}

\begin{document}
%
\title{Learning Shared Dynamics with Meta-World Models}
\author{Lisheng Wu
\thanks{Equal contribution.}
, Minne Li
\samethanks
, Jun Wang \\
University College London\\
\{lisheng.wu.17,minne.li.17,jun.l.wang\}@ucl.ac.uk
}

\newcommand{\minne}[1]{{\bf \color{red} [[Minne says: #1']]}}
\newcommand{\jun}[1]{{\bf \color{blue} [[Jun says ``#1'']]}}
\newcommand{\lisheng}[1]{{\bf \color{violet} [[Lisheng says ``#1'']]}}

\maketitle
\begin{abstract}

Humans have consciousness as the ability to perceive events and objects: a mental model of the world developed from the most impoverished of visual stimuli,
enabling humans to make rapid decisions and take actions.
Although spatial and temporal aspects of different scenes are generally diverse,
the underlying physics among environments still work the same way,
thus learning an abstract description of shared physical dynamics helps human to understand the world.
In this paper, we explore building this mental world with neural network models through multi-task learning, namely the meta-world model.
We show through extensive experiments that our proposed meta-world models successfully capture the common dynamics over the compact representations of visually different environments from Atari Games.
We also demonstrate that agents equipped with our meta-world model possess the ability of visual self-recognition,
i.e., 
recognize themselves from the reflected mirrored environment derived from the classic mirror self-recognition test (MSR).
\end{abstract}

\section{Introduction}
Building machines to mimic human brain's abilities has drawn considerate attention from the recent progress in artificial intelligence (AI) \cite{lake2014towards,Lake1332,GravesHybrid2016,BakerRational17}.
Founders of the modern computational science consider the possibility that machines would ultimately possess consciousness \cite{doi:10.1093/mind/LIX.236.433},
which is the ability to perceive events and objects as humans have \cite{sep-consciousness}.
However, many of the current advances lie on the statistical \emph{pattern recognition} paradigm,
which treats learning as making good predictions by discovering patterns correlated to high value rewards (or low errors) directly from the environments \cite{lake_ullman_tenenbaum_gershman_2017}.
These approaches do not draw inspiration from human cognition aspects, i.e., learning and thinking like a person \cite{Dehaene486}, and are considered as \emph{model-free}.
At the early stage of learning in high-dimensional environments with sparse rewards,
model-free methods cannot find an optimal learning direction due to the lack of reward signals,
thus requiring large amounts of data to explore and learn a good policy.

In contrast, humans can achieve comparable performances on a range of tasks with much less experiences.
For example, a human player needs only two hours of practice before achieving reasonable performance on one Atari game and can quickly adapt to different games,
while DQN needs several days' of training using large amounts of computational resources \cite{mnih2015human}. 

 

Humans develop a mental model of the world from the most impoverished of visual stimuli \cite{10.2307/1416950},
and use this world to make rapid decisions and take actions \cite{Forrester1971}.
To build truly human-like AI machines,
we consider an engineering approach in the first step,
and focus on developing the world model to support explanation and understanding.
One of the key ingredients for building this model is the cognitive capability of understanding the underlying physics and dynamics of the environment \cite{doi:10.1146/annurev.ps.43.020192.002005}.
For example, in the Atari Pong game, the ball and paddles follow principles of persistence, continuity, cohesion and solidity \cite{Bellemare:2015:ALE:2832747.2832830}.
Through mental models of the world, humans can reconstruct a perceptual scene following these principles 
to support mental simulations that can predict the future movement of the objects \cite{SPELKE199029}.
Equipped with a world model understanding these intuitive theories of physics,
agents can simulate the real experiences and learn the structured properties of the environment.
By exploiting the underlying dynamics learned by the model,
we can reduce the dimension of the input and generalize features across states and actions in high-dimensional environments \cite{watter2015embed,wahlstrom2015pixels,levine2014learning}.
With one transition model, agents can attend to the dynamics of the states by modeling how the environment evolves with specific action.
These approaches, regarded as \emph{model-based} learning, have been investigated by several previous works \cite{sutton1990integrated,levine2014learning,watter2015embed,wahlstrom2015pixels,schmidhuber2015learning,gu2016continuous,leibfried2016deep,ha2018world}.
However, the learned underlying dynamics are often restricted to be effective in a single world environment.

Although spatial and temporal aspects of different scenes are generally diverse,
the underlying physics among environments still work the same way.
As such, humans can easily adapt to different environments with the help of their understanding of the underlying physics.
For example, humans can still play Pong game when the observation is mirrored or transposed (rotated by ninety degrees and mirrored).
On the other hand, recent computational achievements cannot solve such scene understanding problems, e.g., not rotation-invariant.
Human-like world models should understand these shared physical dynamics and use them to rapidly generalize knowledge to new tasks and environments.
This common knowledge serves as a prior for learning new task, helping agents to efficiently adapt to new environments.
For example, the knowledge gained from Atari Pong can be beneficial for learning to play Breakout and Video Pinball, which share the common concept of the ball and paddles \cite{parisotto2015actor}.
Multi-task learning (or meta-learning) \cite{duan2016rl,finn2017model,parisotto2015actor} has been used
to learn the common knowledge among different tasks.
However, most meta-learning algorithms generally have troubles in searching common parameters or feature spaces directly from high-dimensional environments.

In this paper, we explore building meta mental world models to establish the common physical dynamics over the 
compact 
representations from visually different environments. 
Without the guidance of explicit rewards, our meta-world models learn about the relationships among different worlds through multi-task learning.
The key concept is to maintain a recurrent neural network (RNN) using self-supervised learning to capture the underlying dynamics of each environment and find a common latent structure across different environments.
As a result, we obtain a low-dimensional common latent structure among multiple environments which share the underlying dynamics learned by the RNN.
We demonstrate the performance of our meta-world using five variants of the Atari Pong game, which are completely different from the original Pong in the transition function and even the state space.

As the only premise we have is that all training environments share the same physical dynamics, two challenges are induced to build meta-world models. 
Firstly, the transformation pattern and corresponding states between environments are unknown, so we cannot simply obtain the common representations by direct transformation or supervised learning. 
Secondly, it's not guaranteed to learn shared dynamics as the capacity of neural networks can be large enough to regard all training environments' dynamics as different even with a single model.
Through extensive experiments, we find that it's possible to unify two visually different environment using the shared dynamics.
We also demonstrate that agents equipped with our meta-world model possess the ability of visual self-recognition, i.e., pass the classic mirror self-recognition test (MSR) \cite{Gallup86}.
To the best of our knowledge, this is the first work to build a meta-world model to learn shared dynamics.

\section{Related Work}
Model-based deep reinforcement learning algorithms have been shown to be more effective than model-free alternatives in certain tasks \cite{watter2015embed,wahlstrom2015pixels,levine2014learning}.
One of the classical model-based algorithms is Dyna-Q \cite{sutton1990integrated}
which learns the policy from both the model and the environment 
by supplementing real world on-policy experiences with simulated trajectories.
However, 
using trajectories from a non-optimal or biased model can lead to learning a poor policy \cite{gu2016continuous}. 
To model the world environment of Atari Games, autoencoder has been used to predict the next observation and environment rewards \cite{leibfried2016deep}.
Some previous works \cite{schmidhuber2015learning,ha2018world,leibfried2016deep} maintain a recurrent architecture to model the world using unsupervised learning
and proved its efficiency in helping RL agents to outperform previous methods in complex environments.
The work from \cite{ha2018world} also demonstrates their model's capability of helping the agent to act in real world by learning from the "dream world", i.e., the mental model of the world. 
However, these models can only be applied to a single environment and need to be built from scratch for new environments.
Although using a similar recurrent architecture, our work differs from above works by learning the common underlying dynamics over multiple environments.

To achieve multi-task learning, recurrent architecture \cite{duan2016rl,wang2016learning} has also been used to learn to reinforcement learn by adapting to different MDPs automatically,
which is shown to be comparable to the UCB1 algorithm \cite{Auer2002} on bandit problems.
Meta-learning shared hierarchies (MLSH) \cite{frans2017meta} shares sub-policies among different tasks to achieve the goal in the training process, where high hierarchy actions are obtained and reused in other tasks. 
Model-agnostic meta-learning algorithm (MAML) \cite{finn2017model} minimizes the total error across multiple tasks by locally conducting few-shot learning to find the optimal parameters
for both supervised learning and RL.
Actor-mimic \cite{parisotto2015actor} distills multiple pre-trained DQNs on different tasks into one single network to accelerate the learning process by initializing the learning model with learned parameters of the distilled network.
To achieve promising results, these pre-trained DQNs have to be expert policies.
Distral \cite{teh2017distral} learns multiple tasks jointly and trains a shared policy as the "centroid" by distillation.
Most of the meta-learning approaches consider the problems within the model-free RL paradigm and focus on finding the common structure in the policy space.
However, learning in high-dimensional environments  with model-free approaches suffers from the sparse and high-variance reward signals, thus requiring large amounts of data to explore.
In contrast, 
we explicitly maintain a meta-world model to capture the latent structures and dynamics of the environment, thus having more stable correlation signals.

In terms of the meta-learning for model-based algorithms, \cite{al2017continuous,clavera2018learning} focused on model adaptation when the model is incomplete or the underlying MDPs are evolving. 
By taking the unlearned model as a new task and continuously learning new structures, the agent can keep its model up to date.
Different from above approaches, we focus on how to 
establish the common physical dynamics over the compact representations from visually different environments.


\section{Preliminaries}
In this section, we will introduce the notation and formalize the meta-world learning.

\subsubsection{Environment Setting}
We first define the environment as a Markov Decision Process (MDP) represented by tuple 
$\Gamma={\langle} \mathcal{S}, \mathcal{P}, \mathcal{A}, \mathcal{R} {\rangle}$,
where $\mathcal{S}, \mathcal{P}, \mathcal{A}, \mathcal{R}$
are the state space, the transition probability function, the action space, and the reward function respectively.
At each time step $t$, a state $s^t \in \mathcal{S}$ is provided by the environment $\Gamma$.
The agent in this environment receives this state information and chooses an action $a^t \in \mathcal{A}$ according to its own strategy.
The environment then receives this action and moves to a new state $s^{t+1}$ following the transition function $\mathcal{P}(s^{t+1}|s^t,a^t): \mathcal{S} \times \mathcal{A} \rightarrow \mathcal{S}$.
The initial state of the environment is determined by a distribution $p_1(s^1): \mathcal{S} \rightarrow [0,1]$.
In the following sections, we also use the apostrophe to indicate quantities at next time step when $t$ is neglected for simplicity, e.g., $s'$ instead of $s^{t+1}$.

To help agents learn compact representations of the environment, \cite{ha2018world} introduced the world model.
Inspired by the human cognitive system,
the world model ($f_\phi$, $g_\theta$) consists of two components: a Vision Model (V) and a Memory Model (M),
denoted as $f_\phi$ and $g_\theta$, parameterized by $\phi$ and $\theta$ respectively.
At each time step $t$, the vision model receives the real-time high-dimensional observation (e.g., a image frame) from the environment and compresses the input into a compact but informative representation $z^t$.
The memory model serves as the history encoder and future predictor by:
(1) compressing the abstract presentation from V and producing a history information $h^t$ over time;
(2) combining the history information $h^{t-1}$ with the current abstract observation $z^t$ to predict the future $z^{t+1}$.
Because the agent affects the environment state with its own action,
agent action $a^t$ is also fed into the memory model to better predict the future.
With the ability of reconstructing low-dimensional representations to the form of environment observations, 
a variational autoencoder (VAE) \cite{kingma2013auto} is normally used as the vision model,
where the encoder and the decoder are denoted as $f^e$ and $f^d$, parameterized by $\phi^e$ and $\phi^d$ respectively.
Agents can then use the information from $z^t$ and $h^t$ to make a decision $a^t$, where
\begin{align}
\label{eq:model_func} 
 & z^t = f^e_\phi(s^t), \\
 & \hat{s}^t  = f^d_\phi(z^t) = f_\phi(s^t), \\
  & \tilde{z}^{t+1}, h^{t+1}  = g_\theta(z^t, h^t, a^t), \\
  & \tilde{s}^{t+1} = f^d_\phi(\tilde{z}^{t+1}).
\end{align}
\subsubsection{Meta-World Model}
To train a world model that can seize the common underlying dynamics of multiple environments $\Gamma_i \in \{\Gamma_1,...,\Gamma_N\}$,
we propose the meta-world model and formalize the problem of meta-world learning.
The goal of meta-world learning is to build one memory model M 
for sharing across multiple environments.
During the training stage, a set of trajectories $D_i=\{(s^1, a^1, ..., s^{T-1}, a^{T-1}, s^T)\}$ are sampled from the environment $\Gamma_i$.
We define the meta-world model as $(f_*, g_\theta)$ 
and the world model for environment $\Gamma_i$ as $(f_{\phi_i}, g_\theta)$, abbreviated as $(f_i, g_\theta)$.
For each environment $\Gamma_i$,
the V model with parameter $\phi_i$ encodes input states $s_i$ to latent vectors $z_i$,
which represents the dynamics learned by the meta-world model.
Our expectation is to learn similar $z_i$,
i.e., shared underlying dynamics across multiple environments.


\section{Methods}
In this section, we first provide the meta-learning algorithm for meta-world models.
Then we present the method to validate the performance of meta-world models.
To find the underlying dynamics of different environments,
we describe some further constraints applied to the intermediate output distribution of the V model.


\begin{figure}[t]
    \centering
    \begin{subfigure}[t]{0.19\textwidth}
       \centering
        \includegraphics[width=\textwidth]{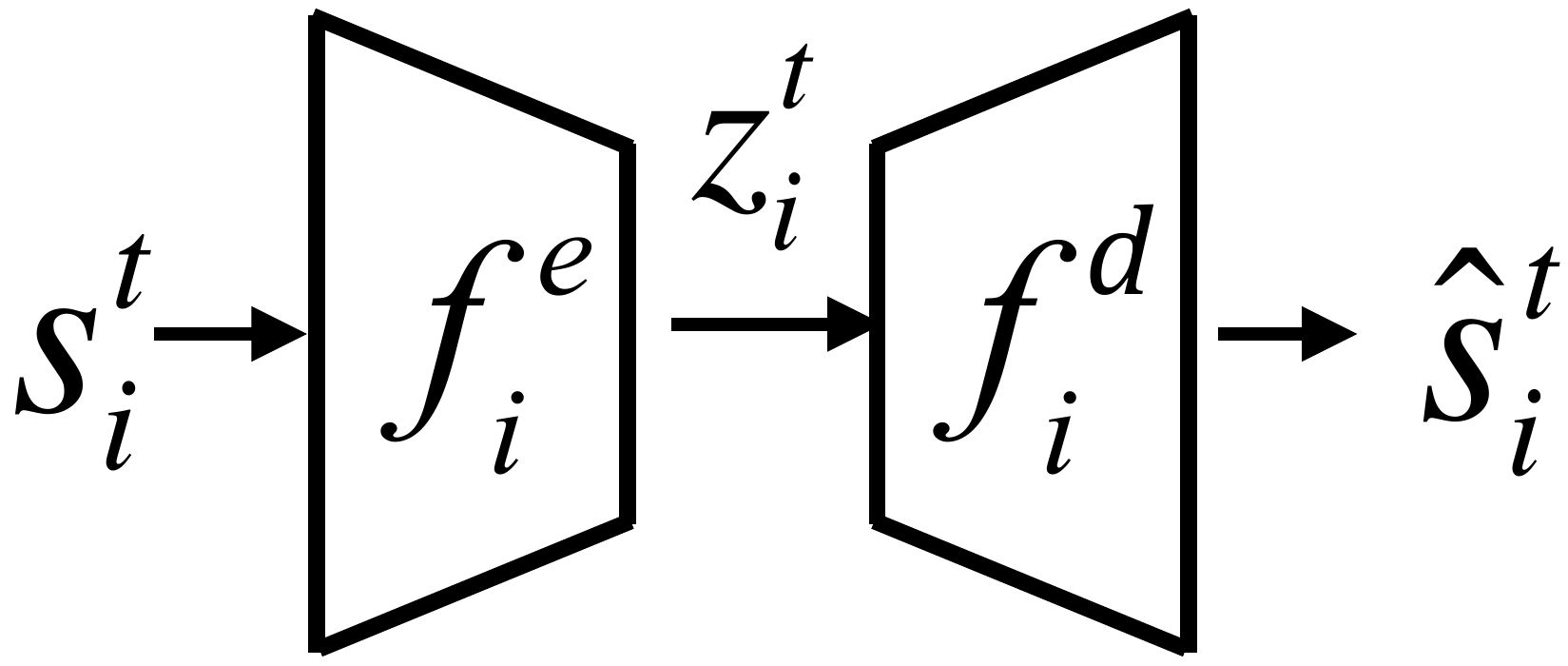}
        \caption{Reconstruction}
    \label{fig:rec_process}
    \end{subfigure}
    \begin{subfigure}[t]{0.25\textwidth}
       \centering
        \includegraphics[width=\textwidth]{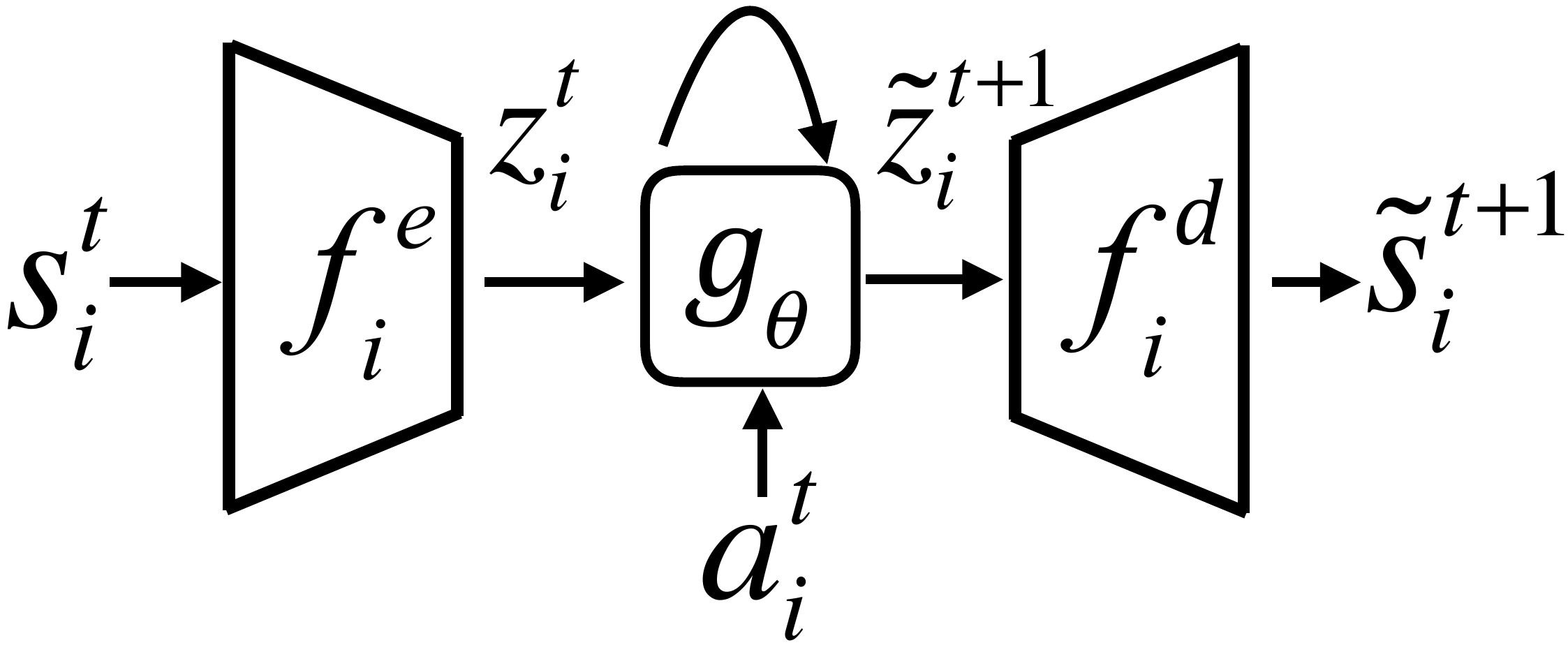}
        \caption{Prediction}
    \label{fig:pred_process}
    \end{subfigure}
    \caption{Meta-learning for meta-world models.}
\label{fig:world}
\end{figure}

\subsection{Meta-Learning for Meta-World Models}
\label{ml_for_mw}
To train a meta-world model, we adopt the variational autoencoder (VAE) as the vision model V and the LSTM \cite{Hochreiter:1997:LSM:1246443.1246450} model with variant output as the memory model M. 

As illustrate in Fig. \ref{fig:world}, the training process consists of two stages: the reconstruction of V models and the prediction of M models.
The reconstruction stage is for V model to compress the input state $s^t$ to the latent vector $z^t$, and uncompress $z^t$ to the output state $\hat{s}^t$ that closely matches the original $s^t$.
In the prediction stage,
the M model takes both the action $a^t$ and the latent vector $z^t$ as inputs to predict $\tilde{z}^{t+1}$.
Then, the predicted state $\tilde{s}^{t+1}$ is obtained by decoding $\tilde{z}^{t+1}$ using the decoder of the V model to approximate the true next state $s^{t+1}$,
which is the result of applying $a^t$ to the environment under the current state $s^t$. 
The reconstruction loss $L_r$ is the distance between the state approximation $\hat{s}^t$ and the true environment state $s^t$, and the prediction loss $L_p$ comes from the distance between the predicted state $\tilde{s}^{t+1}$ and the actual state of the next time step $s^{t+1}$:
\begin{align}
\label{eq:model_loss} 
 \mathcal{L}_{r}(\Gamma_i,\phi_i)  &= \sum\limits_{D_i}\sum\limits_{t} d(\hat{s}^t, s^t),\\
 \mathcal{L}_{p}(\Gamma_i, \theta,\phi_i) & = \sum\limits_{D_i}\sum\limits_{t}d(\tilde{s}^{t+1}, s^{t+1}),
\end{align}
where $d$ is a distance measurement function.
During the training process, the updating rules for both models are defined as
\begin{align}
  \label{eq:model_update}
  {\theta} &\leftarrow \theta-\alpha \sum\limits_{\Gamma_i \sim p(\Gamma)}\nabla_{\theta} \mathcal{L}_{p}(\Gamma_i, \theta,\phi_i), \\
  {\phi_i} &\leftarrow \phi_i-\beta_r \nabla_{\phi_i} \mathcal{L}_{r}(\Gamma_i, \phi_i)-\beta_p \nabla_{\phi_i} \mathcal{L}_{p}(\Gamma_i, \theta,\phi_i),
\end{align}
where $\beta_r$ and $\beta_p$ are the ratio of $L_r$ and $L_p$ at one training step respectively.
The V model is updated with gradients of both $L_r$ and $L_p$, and the M model is updated with gradients of $L_p$ only.

In the meta-learning for meta-world models, we alternate the training process for V models between minimizing $L_r$ and $L_p$ by adjusting $\beta_r$ and $\beta_p$ accordingly.
For the training of V models, minimizing $L_p$ 
helps to adjust V models of different environments with respect to the shared underlying dynamics learned by the memory model M.
However, the adjustment of V models to satisfy the learning of shared dynamics will in turn affect the reconstruction ability of V models.
The optimization of $L_r$ and $L_p$ are influenced by each other similar to a general-sum game, where a equilibrium is hard to reach when two optimization process run simultaneously.
Thus, we use the alternate training scheme as minimizing $L_r$ and $L_p$ for V models respectively throughout the training process to avoid trapping into local minimum.
To capture the shared underlying dynamics across multiple environments, following constraints are considered in our model:
\begin{enumerate}
    \item The Memory model M should make use of the abstract representation of the current state and its own memory to predict the next state with specific action.
    \item The Vision model V should be able to reconstruct the original state from its abstract representation. 
\end{enumerate}


\subsection{Validating the Model Performance}
People may doubt that the meta-world might not learn the common dynamics because a neural network with high capacity could represent multiple dynamics simultaneously.
Despite the strict constraints 
listed above, it's still possible for our meta-world model to learn multiple dynamics simultaneously or be trapped into local minimum.
In this section, we present the method to validate that meta-world models actually learn the common dynamics of multiple environments.

Suppose we have environment $\Gamma_i$ and environment $\Gamma_j$ sharing the same underlying physical dynamics but different visual observations.
For the sake of simplicity, we assume the visual observation of $\Gamma_i$ is the transpose of that of $\Gamma_j$, i.e., clockwise rotated by $90^\circ$ and horizontal flipped.
For most of the advanced neural networks without rotation-invariance, these two environments are totally different. 
Although they might be seen as different to humans at the first glance,
humans will come to realize $\Gamma_i$ and $\Gamma_j$ are almost identical because they consist of the same components and share the same underlying physical dynamics,
even though the dynamics are presented in different directions.
Thus, humans have the ability to unify different observations sharing the same underlying dynamics. 

To demonstrate that our meta-world model is also capable of capturing the common dynamics of such two environments instead of treating them as different dynamics and absorbing them simultaneously into the neuron model,
$\Gamma_i$ and $\Gamma_j$ should have very similar abstract representations $z_i$ and $z_j$ of corresponding observations $s_i$ and $s_j$. 
However, it's not applicable to directly measure the distance of the abstract representations in the vector space, due to the high-variance of V models.
Instead, we validate our model by encoding the observation $s_i$ from $\Gamma_i$ to the corresponding abstract representation $z_i$,
which is decoded with the decoder of the V model of environment $\Gamma_j$.
Then, we compare the decoded result $\hat{s}_j$ with the corresponding observation $s_j$.
If $s_j$ and $\hat{s}_j$ are almost identical, we can confirm that meta-world models actually capture the underlying physical dynamics of different environments and unify their abstract representations as the shared dynamics,
rather than simply memorizing and absorbing them  into the neuron models.

\subsection{Adding Constraints on Abstract Representations}
As described in the above section, we want to unify the abstract representations of different environments sharing the same underlying physical dynamics.
However, the abstract representation generated by each vision model $V_i \in {V_1, ..., V_N}$ might not be the same without further constraints on the learning period.
Concretely, because we adopt the VAE as the V model, the element-wise similarity of the abstract representation $z_i$ is determined by the distribution (both the mean and the variance) of the intermediate output of VAE.
Thus, it is difficult for randomly initialized V models to find the common dynamics through the coordination of abstract representations.

To add further constraints on the distribution of the abstract representations $z_i$,
we introduced the Maximum Mean Discrepancy \cite{gretton2007kernel} to regularize the channels of $z_i$ by adding constraints on the distance among all the mean and variance of the intermediate output of V models.
Concretely, we force all meta-world models to have similar distributions to the abstract representation of $\Gamma_1$ by adding the total training loss with a regularizer
\begin{equation}
\label{eq:model_update}
 \mathcal{L}_{MMD} = \sum\limits_{-\Gamma_1}{\eta_\mu||\bar{\mu}_i - \bar{\mu}_1||^2 + \eta_\sigma||\overline{log(\sigma_i)} - \overline{log(\sigma_1)}||^2},
\end{equation}
where $\mu_i$ and $\sigma_i$ are the mean and variance of the VAE of $\Gamma_i$, and $\eta_\mu$ and $\eta_\sigma$ are the ratio of the error of mean and log variance respectively.
Then the total training loss of the V model can be written as,
\begin{equation}
\label{eq:model_update}
 \mathcal{L} = \beta_p \mathcal{L}_p + \beta_r \mathcal{L}_r + \eta \mathcal{L}_{MMD}.
\end{equation}
During the training process, we add MMD loss to the reconstruction stage, while keeping training with prediction loss in the prediction stage.
In this way, the distributions of abstract representations can also be influenced by other environments through the participation of memory model M.
That is, the V models of other environments ${V_2, ..., V_N}$ can affect M to change its parameters by the joint training in the prediction stage, thus influencing the distributions of abstract representations of $V_1$ indirectly. 

\begin{figure*}[t]
    \centering
    \begin{subfigure}{0.19\textwidth}
        \includegraphics[height=0.48\textwidth]{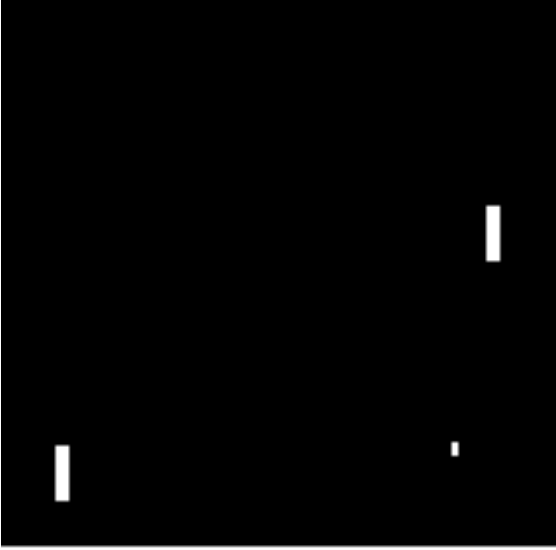}
        \includegraphics[height=0.48\textwidth]{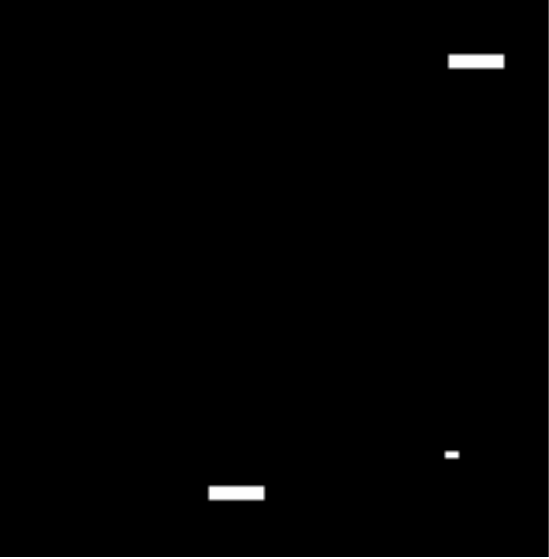}
        \caption{$\Gamma_o \rightarrow \Gamma_t$}
    \label{fig:transpose world}
    \end{subfigure}
    \begin{subfigure}{0.19\textwidth}
        \includegraphics[height=0.48\textwidth]{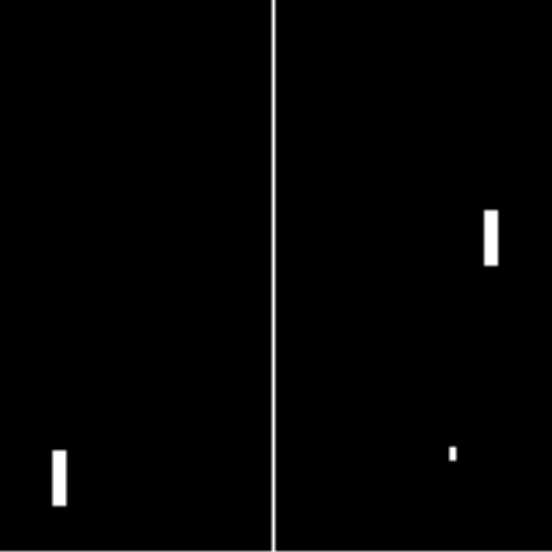}
        \includegraphics[height=0.48\textwidth]{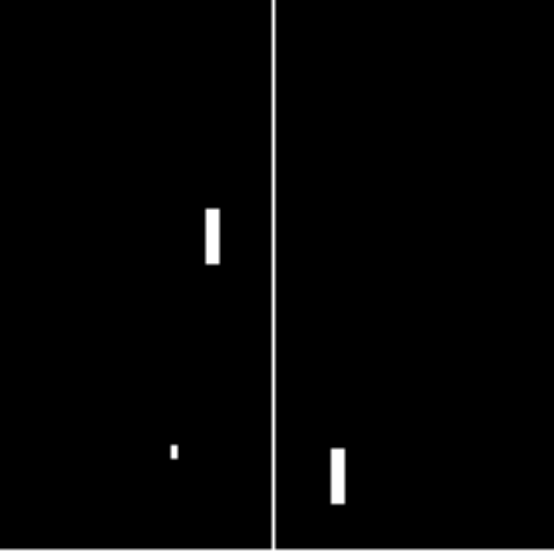}
        \caption{$\Gamma_o \rightarrow \Gamma_h$}
    \label{fig:split1 world}
    \end{subfigure}
    \begin{subfigure}{0.19\textwidth}
        \includegraphics[height=0.48\textwidth]{figures/env/origin.pdf}
        \includegraphics[height=0.48\textwidth]{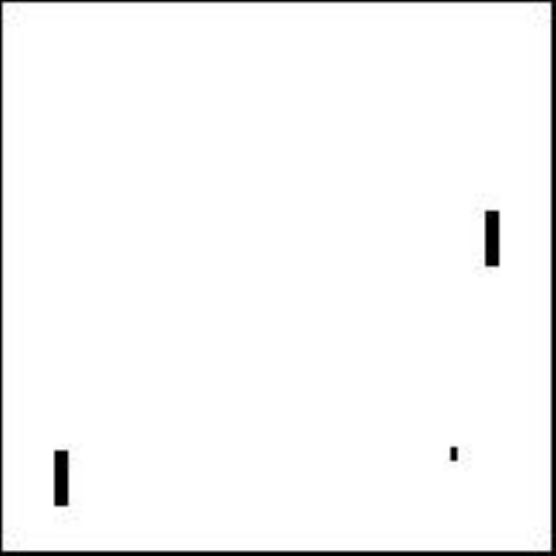}
        \caption{$\Gamma_o \rightarrow \Gamma_c$}
    \label{fig:color world}
    \end{subfigure}
    \begin{subfigure}{0.19\textwidth}
        \includegraphics[height=0.48\textwidth]{figures/env/origin.pdf}
        \includegraphics[height=0.48\textwidth]{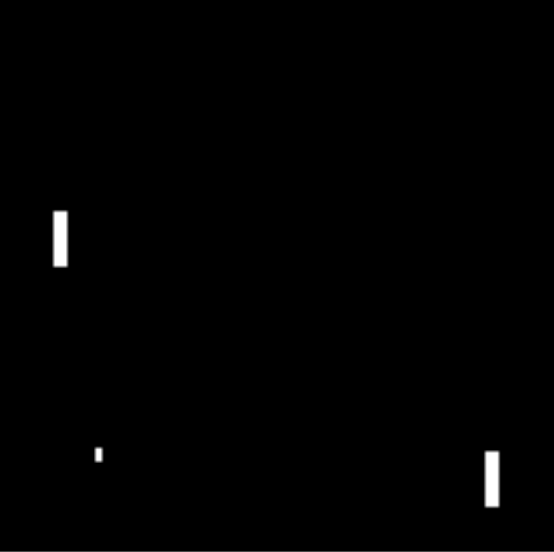}
        \caption{$\Gamma_o \rightarrow \Gamma_m$}
    \label{fig:mirror world}
    \end{subfigure}
    \begin{subfigure}{0.19\textwidth}
        \includegraphics[height=0.48\textwidth]{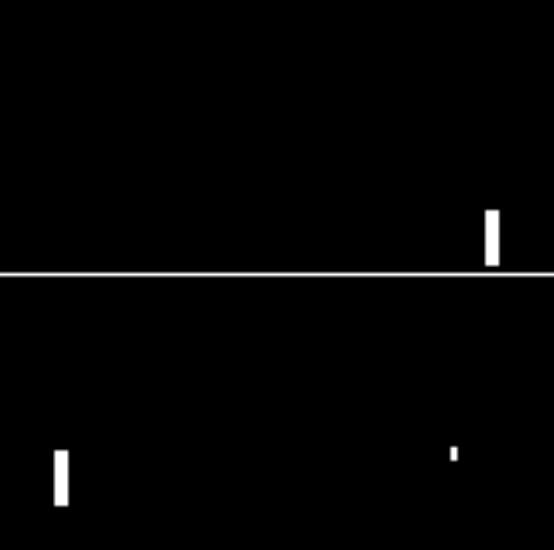}
        \includegraphics[height=0.48\textwidth]{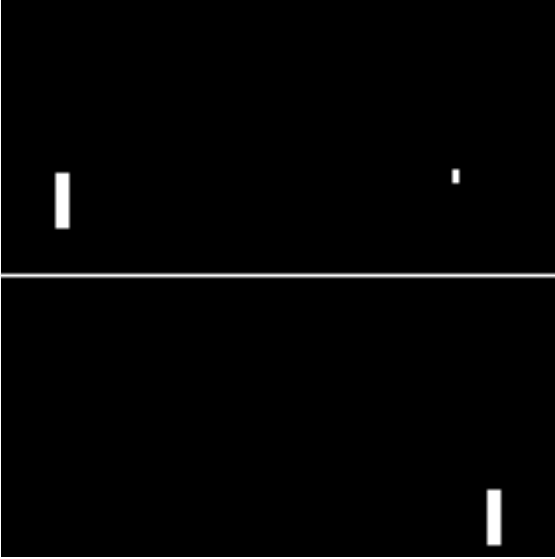}
        \caption{$\Gamma_o \rightarrow \Gamma_v$}
    \label{fig:split2 world}
    \end{subfigure}
    \caption{Meta-world environments.}
    \label{fig:world_trans}
\end{figure*}

%

\section{Experiments}
This section starts with the introduction of our experiment environment (the Atari Pong game), problem setup, and evaluation criteria.
We then present the experiment results and analysis of different training strategies,
i.e., using corresponding states from different environments as training data or not.
We also demonstrate that agents equipped with our meta-world model possess the ability of visual self-recognition as being self-aware.


\subsection{The Pong World}
Atari Pong game is a environment with two paddles that can only move up or down to hit the ball.
The dynamics of Pong game involves an opponent which the player's action cannot control, and an agent that the player can control
To verify the ability of our models capturing shared physical dynamics among different visual observations,
we generate five variants of the Atari Pong game while keeping the physical dynamics of the environment.
As shown in Fig.\ref{fig:world_trans}, each variant corresponds to one transformation from $\Gamma_o$:
(a) the transposed $\Gamma_t$, which is transformed from the state observation of $\Gamma_o$ by clockwise rotating $90^\circ$ and horizontal flipping;
(b) the horizontal-swapped $\Gamma_h$, which is generated by vertically splitting the observation frame of $\Gamma_o$ from the center and swapping the left part with the right part;
(c) the inverse $\Gamma_c$, which is created by exchanging the background color with the paddles/ball color of $\Gamma_o$;
(d) the mirror-symmetric $\Gamma_m$, which reflects $\Gamma_o$ like a mirror by horizontally swapping the observation;
and (e) the vertical-swapped $\Gamma_v$, which is obtained by horizontally splitting the observation of $\Gamma_o$ from the center and swapping the upper part with the lower part.
We should note that the split line in the transformation from $\Gamma_o$ to $\Gamma_v$ may cut the paddles into two halves,
e.g., a part of the paddle could disappear from the top of the frame and reappear at the bottom when the paddle is moving across the center.
We consider this situation as the \emph{paddle teleportation}.

These variants can be divided into two groups by the difference of state space compared to $\Gamma_o$.
The first group, including $\Gamma_t$, $\Gamma_h$, and $\Gamma_c$,
has totally different state space from $\Gamma_o$ by either putting the paddles in different positions or expressing the corresponding states with different pixel values.
Obviously, actions in this group of environments appear differently because of the new state space.
The second group, consisting of $\Gamma_m$ and $\Gamma_v$, has either the same or at least some overlaps with the state space of $\Gamma_o$.
However, the actions still appear differently because of the transformation.

Different from the original Atari Pong observations, we (1) transform each image frame to a binary matrix;
(2) remove the scoreboard to focus on the dynamics of the paddles;
and (3) resize each frame to $64 * 64$ to serve as the state observations of the original Pong world environment $\Gamma_o$.
The action space is formed by all six available discrete actions of the original Atari game environment.

\begin{figure*}[t]
    \centering    
     \begin{subfigure}{0.19\textwidth}
        \includegraphics[width=\textwidth]{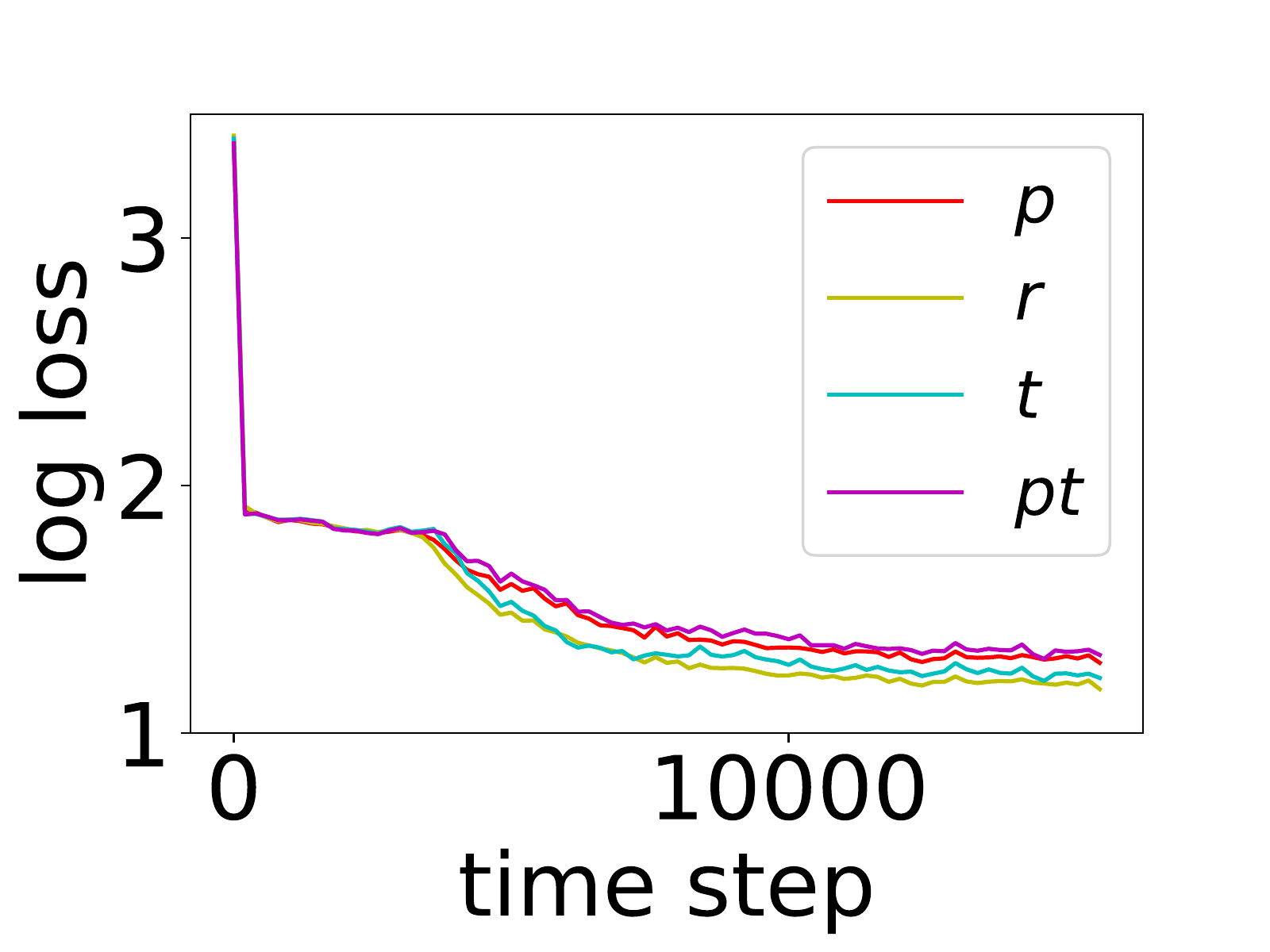}
        \caption{$\Gamma_o$ and $\Gamma_t$}
        \label{fig:cor_transpose_world}
    \end{subfigure}
     \begin{subfigure}[h]{0.19\textwidth}
        \includegraphics[width=\textwidth]{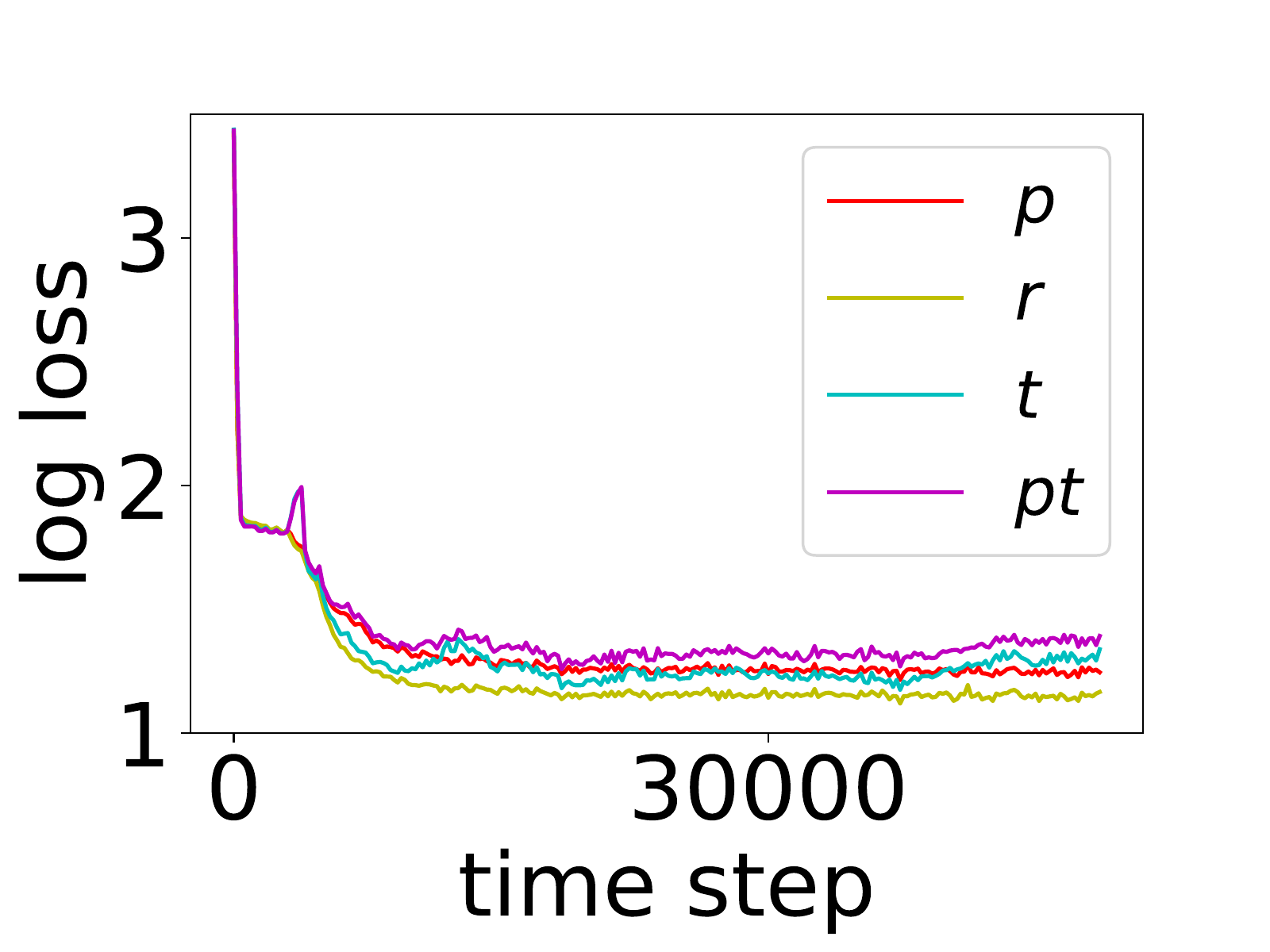}
        \caption{$\Gamma_o$ and $\Gamma_h$}
        \label{fig:cor_concat1_world}
    \end{subfigure}
    \begin{subfigure}[h]{0.19\textwidth}
        \includegraphics[width=\textwidth]{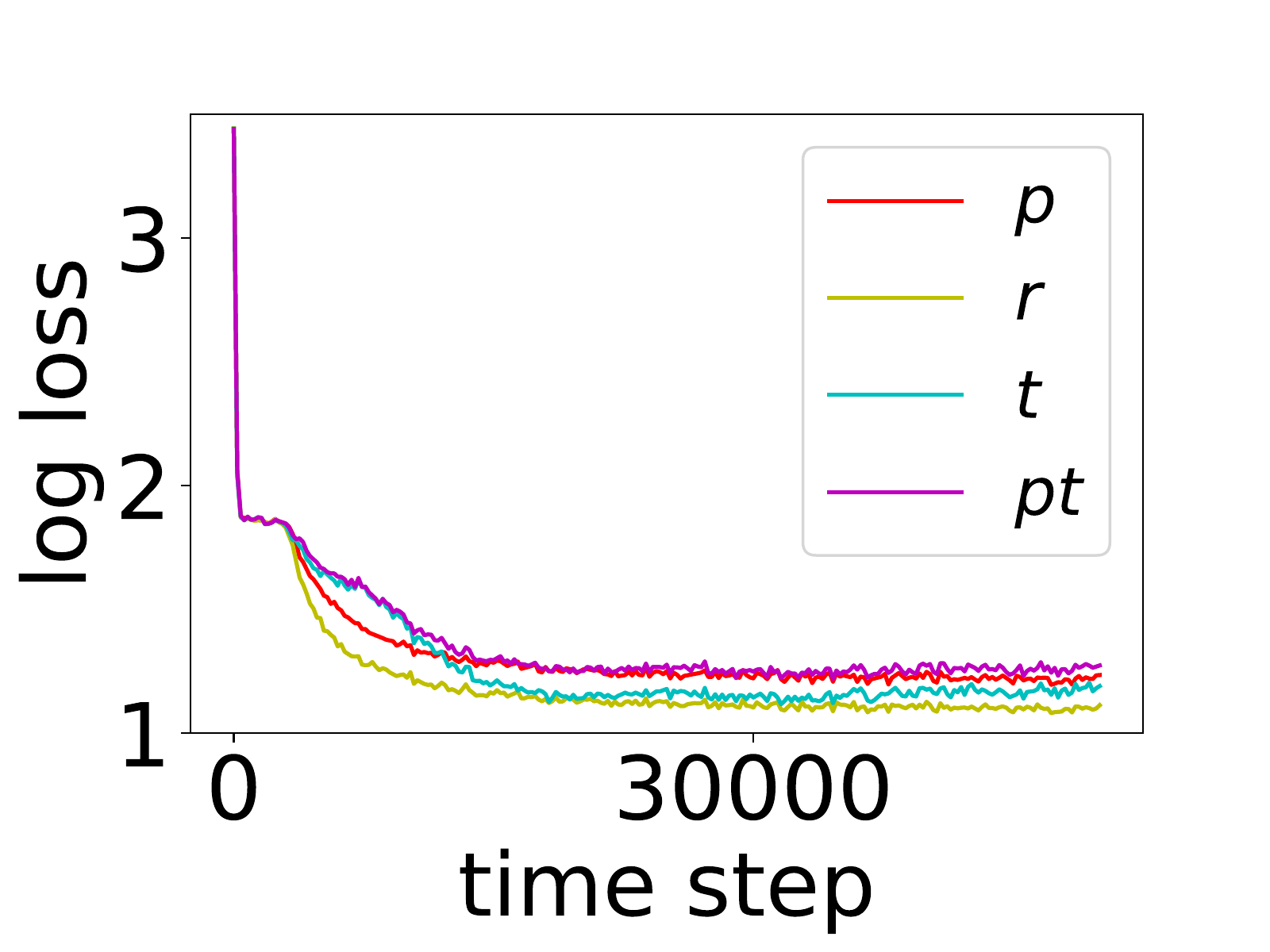}
        \caption{$\Gamma_o$ and $\Gamma_c$}
        \label{fig:cor_color_world}
    \end{subfigure}
    \begin{subfigure}{0.19\textwidth}
        \includegraphics[width=\textwidth]{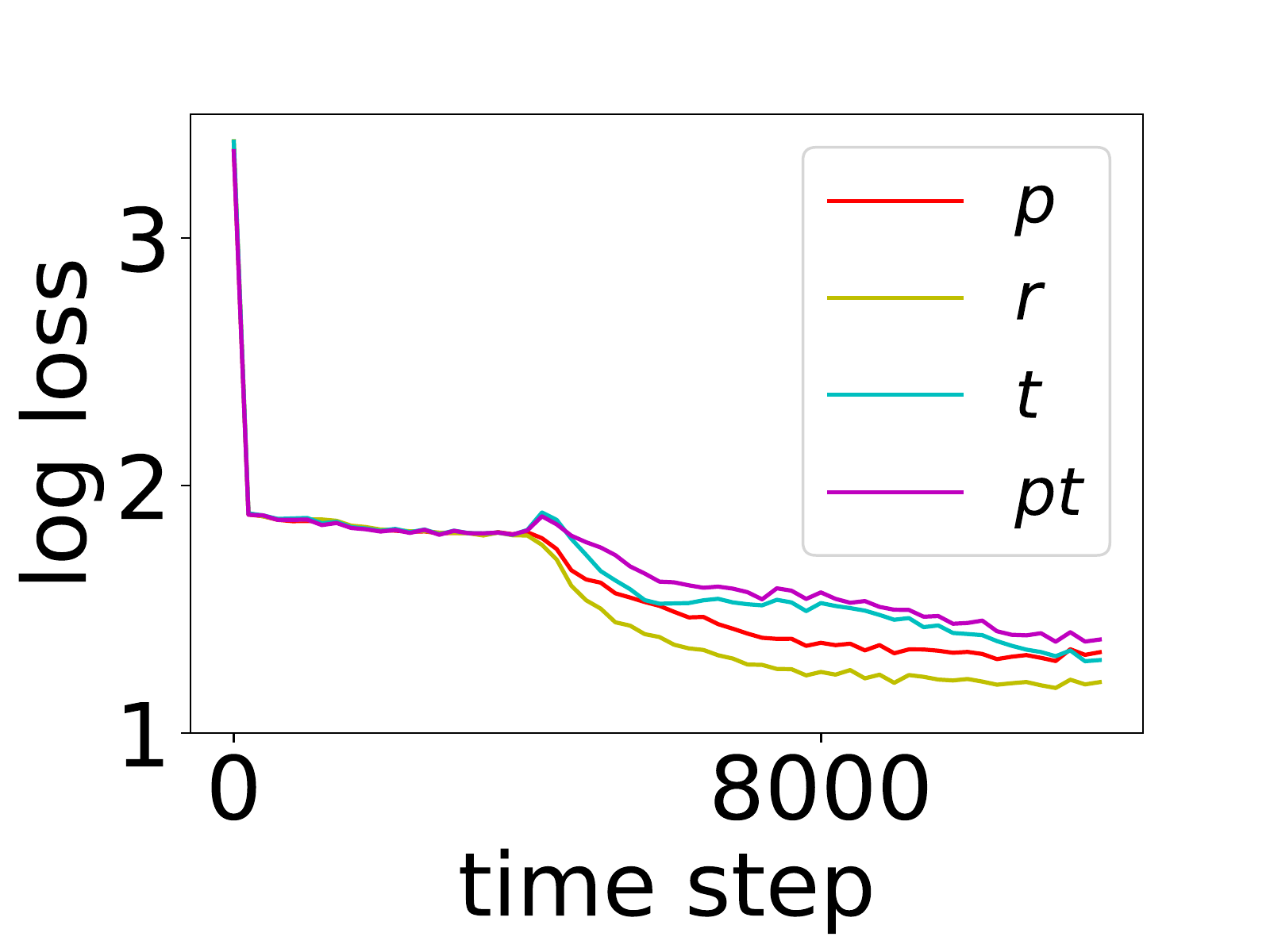}
        \caption{$\Gamma_o$ and $\Gamma_m$}
        \label{fig:cor_mirror_world}
    \end{subfigure}
    \begin{subfigure}[h]{0.19\textwidth}
        \includegraphics[width=\textwidth]{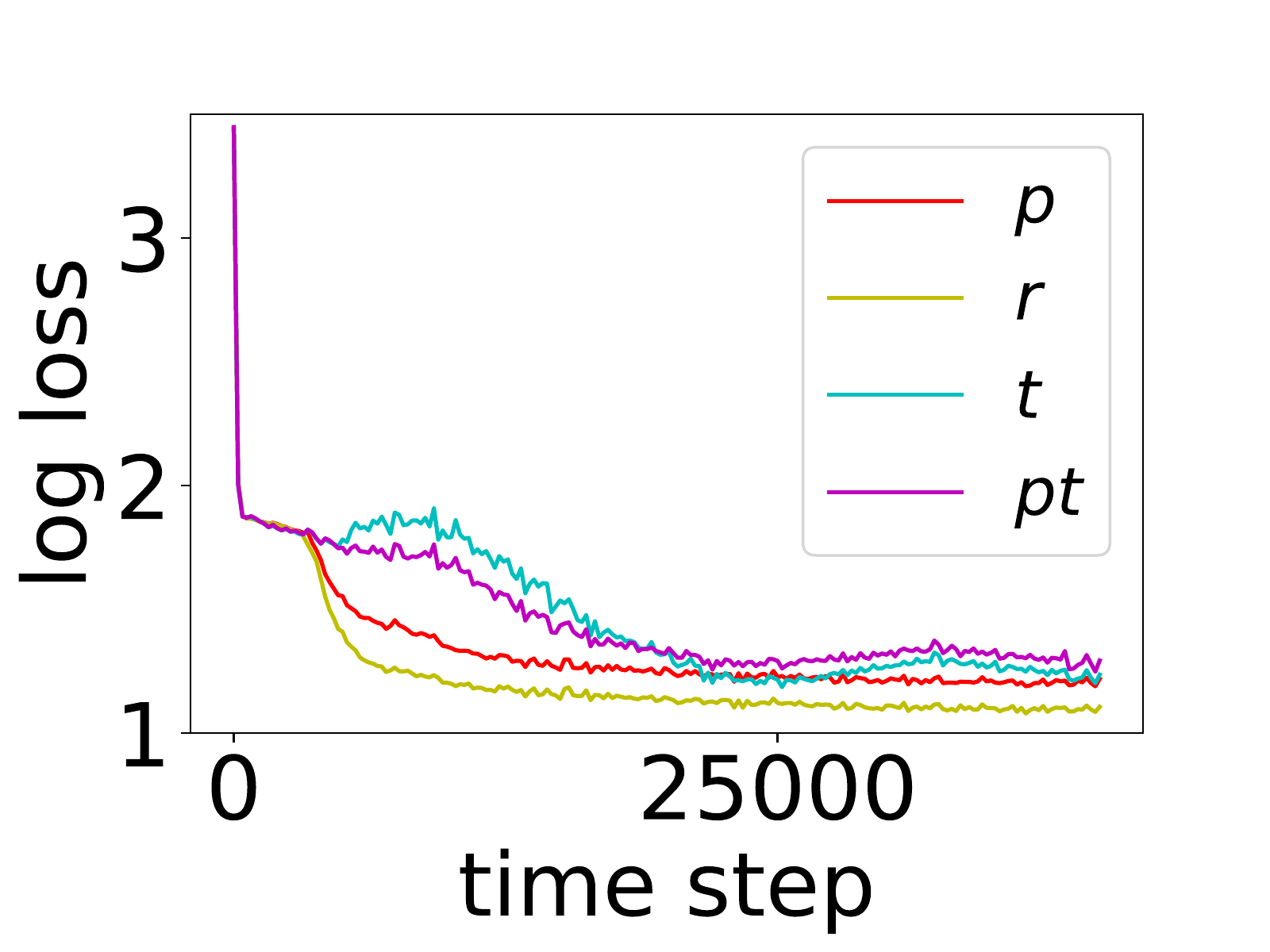}
        \caption{$\Gamma_o$ and $\Gamma_v$}
        \label{fig:cor_concat2_world}
    \end{subfigure}
    \caption{Training meta-world models with corresponding inputs. Legend: $p$: prediction loss $\mathcal{L}_p$; $r$: reconstruction loss $\mathcal{L}_r$; $t$: transformation loss $\mathcal{L}_t$; $pt$: predicted transformation loss $\mathcal{L}_{pt}$.}
    \label{cor_meta_exp}
\end{figure*}

\subsection{Problem Setup}
Although the presented environment variants transit with actions in different ways from the aspect of observations,
they share the same underlying physical dynamics as actions impose transition on the paddle in corresponding ways.
By observing the shared dynamics from environment variants,
humans could recall the dynamics seen in the original environment $\Gamma_o$ and try to find the similarities,
then finally consider them as highly related environments.
However, artificial neural networks (ANNs) are not able to generalize across such transformation with ease.
For example, ANNs are not rotation-invariant,
thus regarding the transposed environment $\Gamma_t$ as a totally different world as $\Gamma_o$.
Our target is to unify the latent representations between environment variants and the original environment by utilizing their shared dynamics.
For simplicity, we build meta-world models on the original Pong environment $\Gamma_o$ and one of the five environment variants separately.
We explore training the shared dynamics on two environments with corresponding inputs and non-corresponding inputs, i.e., using corresponding states from different environments as training data or not, to verify the performance of meta-world models.

To collect the training data covering most dynamics of the Pong environment, we use an agent with random policy to play the game for 10,000 episodes and limit the episode length to 1000 steps.
The vision model V adopts the same architectures as the World Model \cite{ha2018world} with the latent vector $z$ of size 32.
The memory model M is a LSTM with 32 hidden units to predict the parameters of a diagonal
Gaussian distribution.
The predicted latent representation of the next state $\tilde{z}$ is then sampled from the Gaussian distribution.
The latent vector encoded by VAE is sampled from the same Gaussian distribution
, making it enough for an LSTM with one Gaussian prediction to deal with the experiments.
We set the batch size for each task as 16 and the sequence length of LSTM as 25.
As described before, we alternate the training process for V models between minimizing $L_r$ and $L_p$ by setting each training iteration with 20 prediction iterations and 10 reconstruction iterations. 

\subsection{Evaluation Criteria}
Each experiment involves two environments (denoted by $\Gamma_i$ and $\Gamma_j$) as we want to validate whether the latent representations of the corresponding states are identical, or, at least close enough.
Because of the variance introduced by the VAE, we can indirectly decode the latent representation $z_i$ of the state $s_i$ from $\Gamma_i$ by the decoder $f_j^d$ of $\Gamma_j$ to $\hat{s}_j$.
The performance is evaluated by the transformation loss
\begin{equation}
\label{eq:transform_cost}
 \mathcal{L}_t(\Gamma_i, \Gamma_j) = ||s_j - \hat{s}_j||^2,
\end{equation}
which is the distance between $\hat{s}_j$ and the real corresponding states $s_j$.

We also investigate whether the RNN prediction of the latent representation for each environment shares the same vector space.
We decode the predicted latent representation $\tilde{z}'_i$ of environment $\Gamma_i$ using the decoder $f_j^d$ of $\Gamma_j$ to get $\tilde{s}_j'$.
Similarly, the performance is evaluated by the predicted transformation loss
\begin{equation}
\label{eq:predict_transform_cost}
 \mathcal{L}_{pt}(\Gamma_i, \Gamma_j) = ||s_j' - \tilde{s}_j'||^2,
\end{equation}
which is the distance between $s_j'$ and $\tilde{s}_j'$.

\subsection{Meta-World Experiments}
\subsubsection{Corresponding Inputs}
In the first experiment, we train two environments with corresponding states as input.
However, we don't provide meta-world models with any information about the type of the transformation between environments,
thus preventing from simply obtaining the common representations by direct transformation or supervised learning.
For simplicity, we focus on pair-wise training, i.e., fix the original environment $\Gamma_o$ and choose one of the five variants $\Gamma_i \in \{\Gamma_t, \Gamma_h, \Gamma_c, \Gamma_m, \Gamma_v \}$ for each training process.
At each time step, we randomly sample 16 trajectories of length 25 from the dataset as the training data for $\Gamma_o$,
and transform these samples to corresponding states as the training data for $\Gamma_i$,
thus making the training input for different environments already have the same type of dynamics.

We present training results with respect to $\mathcal{L}_p$, $\mathcal{L}_r$, $\mathcal{L}_t$, and $\mathcal{L}_{pt}$ for each of the five environment variant to illustrate the performance of meta-world models.
As shown in Fig. \ref{cor_meta_exp}, the transformation loss $\mathcal{L}_t$ of all experiments could converge to a relatively low value,
which is nearly the same as the reconstruction loss $\mathcal{L}_r$ of the corresponding environment variant;
same results stand for the predicted transformation loss $\mathcal{L}_{pt}$ and prediction loss $\mathcal{L}_p$ as well.
At each training step of the prediction stage, 
the corresponding inputs setting provides meta-world models with same types of dynamics from different environments, which are easier to learn by the memory model.
Meanwhile, vision models could learn to adapt to these shared dynamics simultaneously,
thus leading to easier convergence of meta-world models.
With the help of the memory model predicting corresponding states,
vision models of different environments could understand corresponding states in the same way,
i.e., meta-world models could unify the latent representations of different vision models, thus possessing the ability to capture the underlying shared dynamics of different environments.

\begin{figure*}[htbp]
    \centering    
     \begin{subfigure}{0.19\textwidth}
        \includegraphics[width=\textwidth]{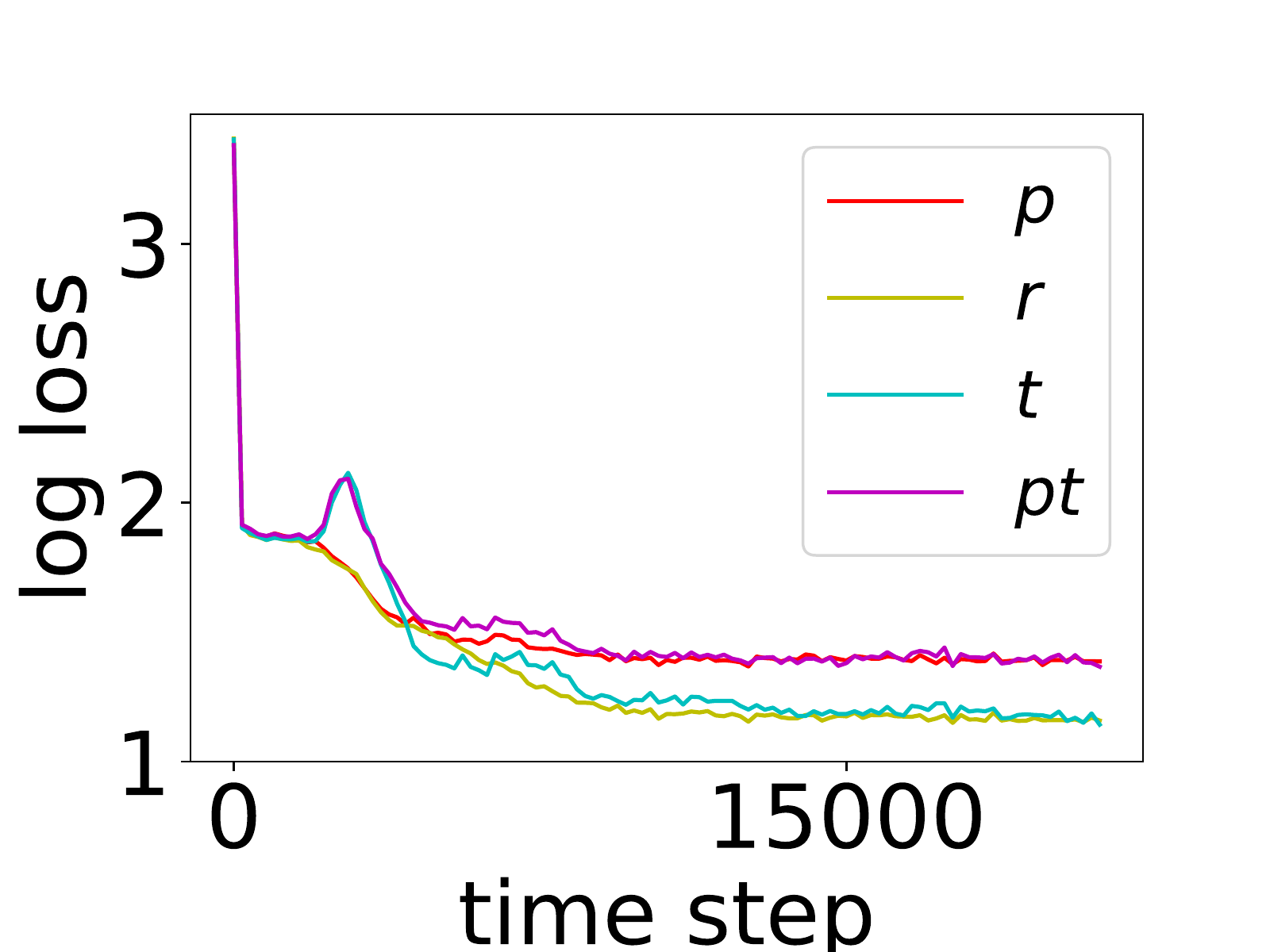}
        \caption{$\Gamma_o$ and $\Gamma_t$}
        \label{fig:transpose_world}
    \end{subfigure}
     \begin{subfigure}[h]{0.19\textwidth}
        \includegraphics[width=\textwidth]{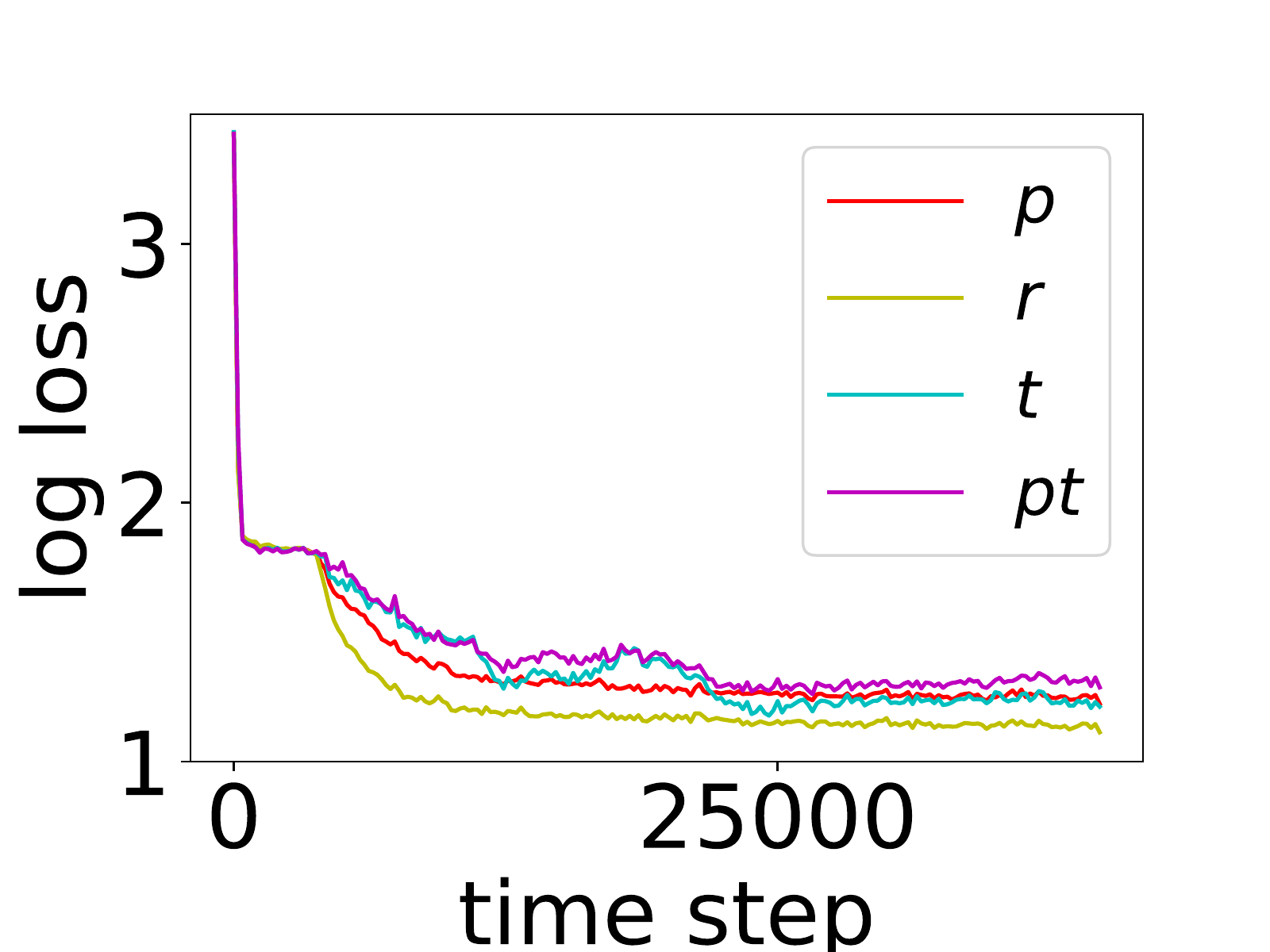}
        \caption{$\Gamma_o$ and $\Gamma_h$}
        \label{fig:concat1_world}
    \end{subfigure}
    \begin{subfigure}[h]{0.19\textwidth}
        \includegraphics[width=\textwidth]{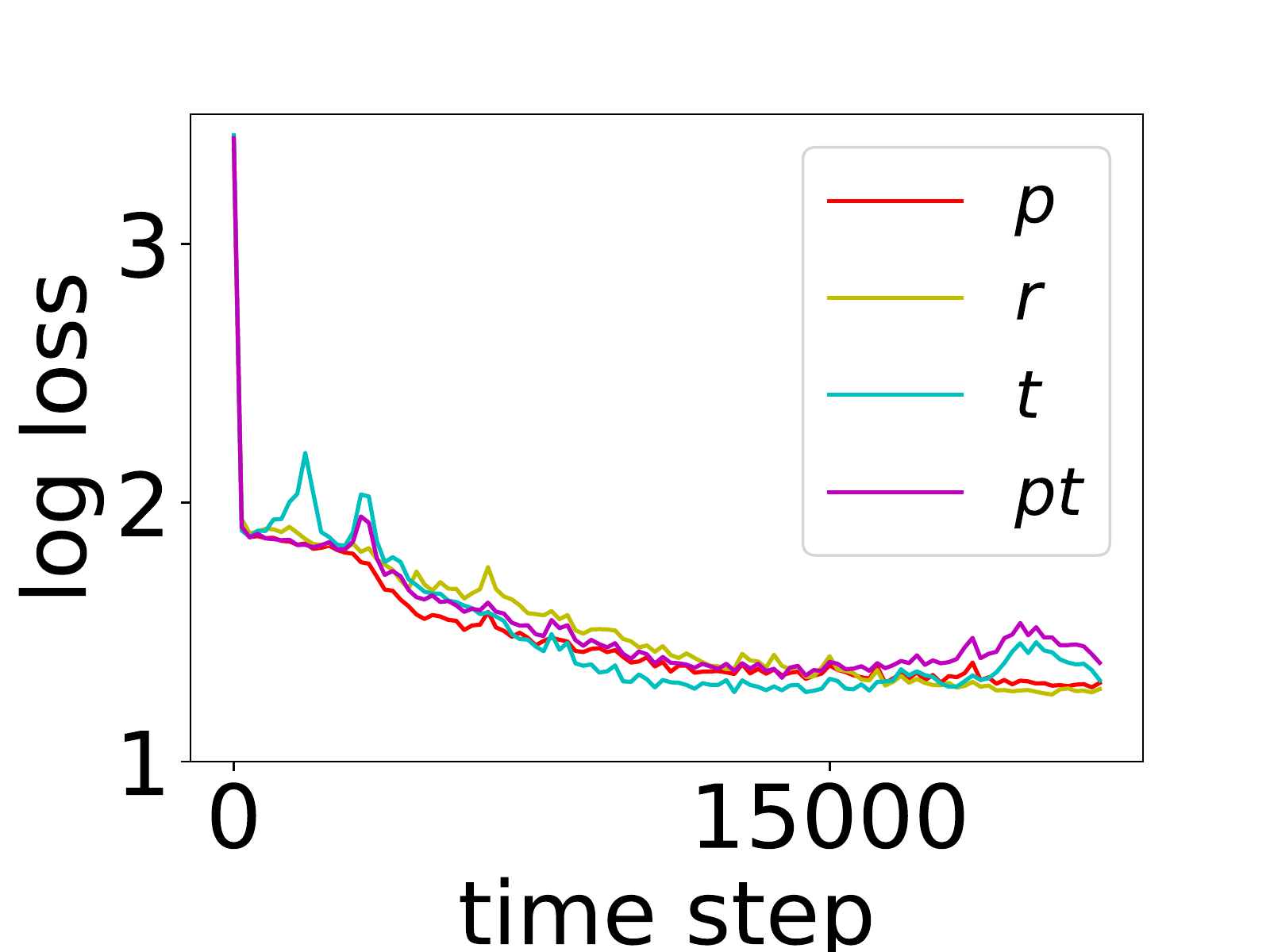}
        \caption{$\Gamma_o$ and $\Gamma_c$}
        \label{fig:color_world}
    \end{subfigure}
    \begin{subfigure}{0.19\textwidth}
        \includegraphics[width=\textwidth]{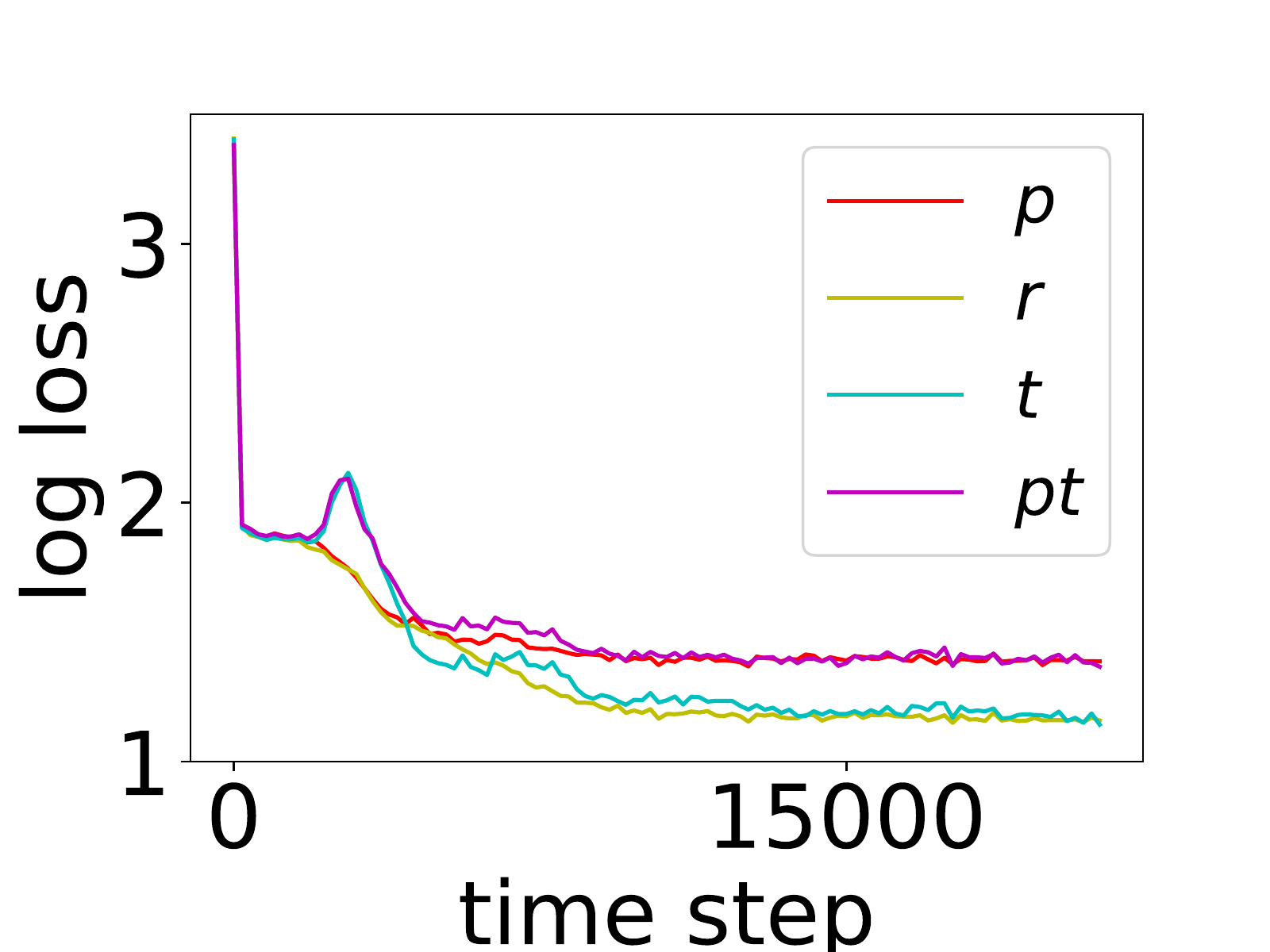}
        \caption{$\Gamma_o$ and $\Gamma_m$}
        \label{fig:mirror_world}
    \end{subfigure}
    \begin{subfigure}[h]{0.19\textwidth}
        \includegraphics[width=\textwidth]{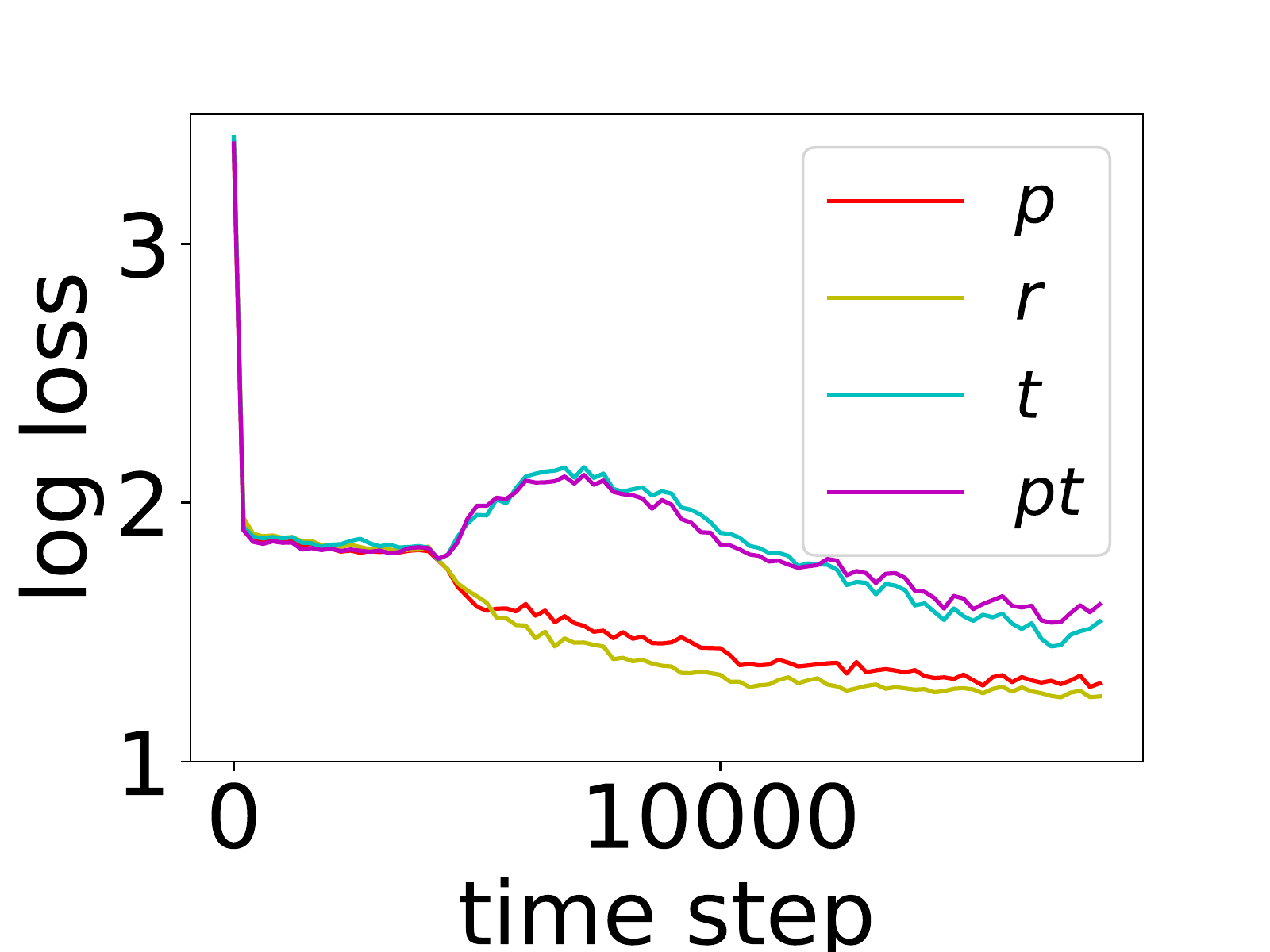}
        \caption{$\Gamma_o$ and $\Gamma_v$}
        \label{fig:concat2_world}
    \end{subfigure}
    \caption{Training meta-world models without corresponding inputs. Legend: $p$: prediction loss $\mathcal{L}_p$; $r$: reconstruction loss $\mathcal{L}_r$; $t$: transformation loss $\mathcal{L}_t$; $pt$: predicted transformation loss $\mathcal{L}_{pt}$.}
    \label{meta_exp}
\end{figure*}

\subsubsection{Non-corresponding Inputs}
In the second experiment, we explore a more humanlike but difficult learning strategy: 
learning shared dynamics without knowing the corresponding states,
i.e., randomly choosing training data instead of providing the model with trajectories with same types of dynamics as input.
Concretely, we split the training set into two parts and conduct the corresponding transformation on one part to prepare the training data for $\Gamma_i$.
At each time step, we randomly sample 16 trajectories from each part and feed them as input to meta-world models.
These trajectories normally present different types of dynamics,
which prevent the memory model from learning corresponding parts of dynamics simultaneously from both environments,
thus requiring models to figure out the exact corresponding states and then learn the shared dynamics.

Similar to experiments with corresponding inputs, we also present training results with respect to $\mathcal{L}_p$, $\mathcal{L}_r$, $\mathcal{L}_t$, and $\mathcal{L}_{pt}$ for the original environment and each of the five variant.
As illustrated in Fig. \ref{meta_exp}, the convergence results on experiments of $\Gamma_o$ and $\Gamma_i \in \{\Gamma_t, \Gamma_h, \Gamma_c, \Gamma_m\}$ prove the effectiveness of meta-world models in unifying the latent representations of different environments when training with non-corresponding inputs, although being slightly worse than that of using corresponding inputs.
This is mainly because vision models also try to adapt to the memory model with previous learned dynamics when the memory model is updated during the prediction stage.
During the reconstruction stage, the memory model is in turn prevented from overfitting because it needs to predict proper latent representation $\tilde{z}'_i$ to get interpreted by the decoder $f^d_i$ of environment $\Gamma_i$.
Thus, meta-world models are able to capture the underlying shared dynamics even though the training process is unstable,
i.e., both the memory model and vision models are optimized through fluctuated directions.

Meanwhile, the meta-world model built on $\Gamma_o$ and $\Gamma_v$ finds it hard to capture the shared dynamics when training with non-corresponding inputs.
We consider it may be attributed to the \emph{paddle teleportation};
in $\Gamma_v$, a paddle can be divided into two parts where one part could disappear at one side of the frame and reappear at the other side,
resulting in new properties of split and teleported paddles not existing in other environments.
Fig. \ref{fig:concat2_world} illustrates the learning curve of a successful try to learn the meta-world model on $\Gamma_o$ and $\Gamma_v$.

\begin{figure}[t]
    \centering
     \begin{subfigure}[h]{0.23\textwidth}
        \includegraphics[width=0.9\textwidth]{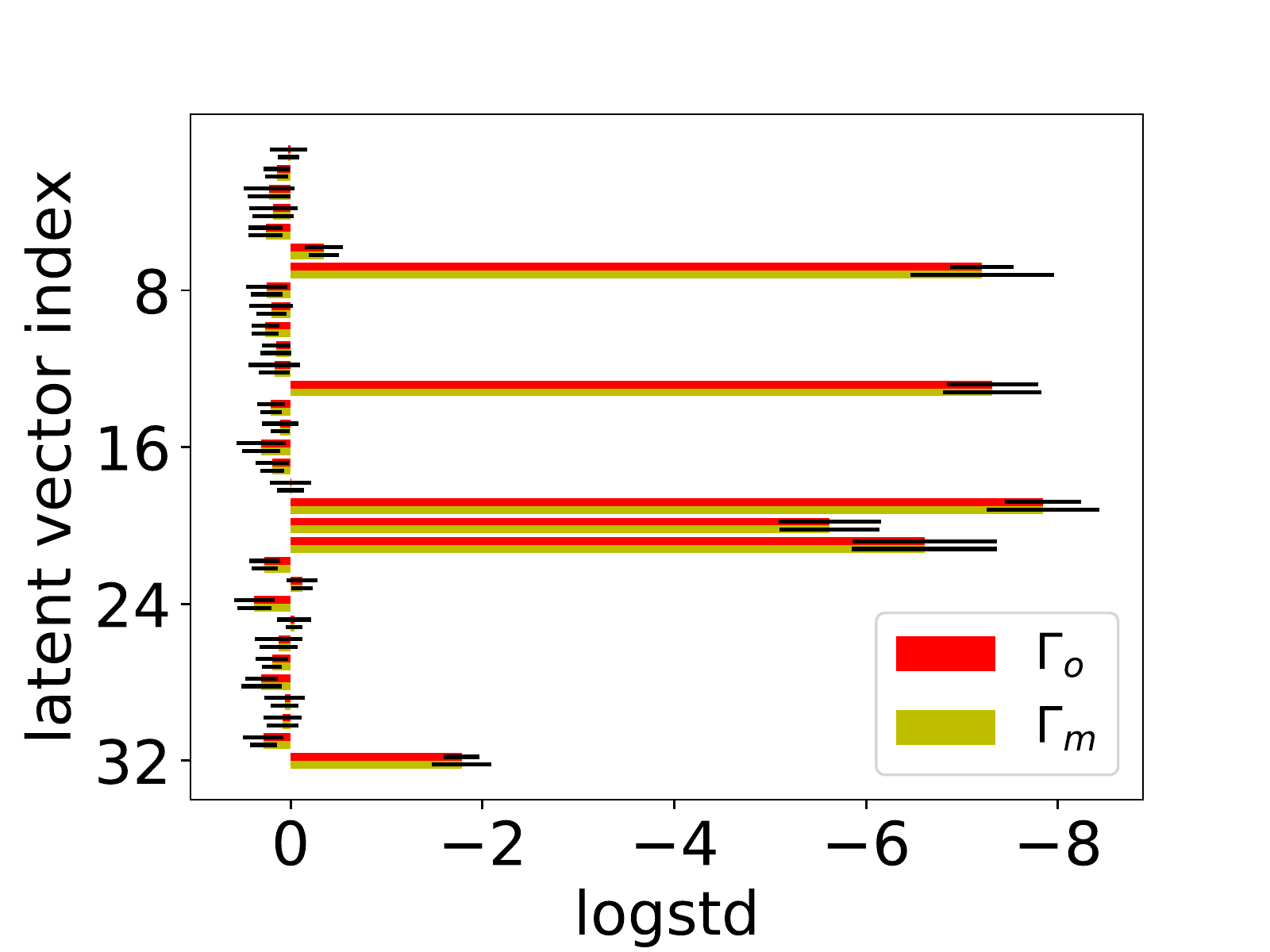}
        \caption{log standard deviations}
        \label{fig:logstd_z}
    \end{subfigure}
      \begin{subfigure}{0.23\textwidth}
        \includegraphics[width=0.9\textwidth]{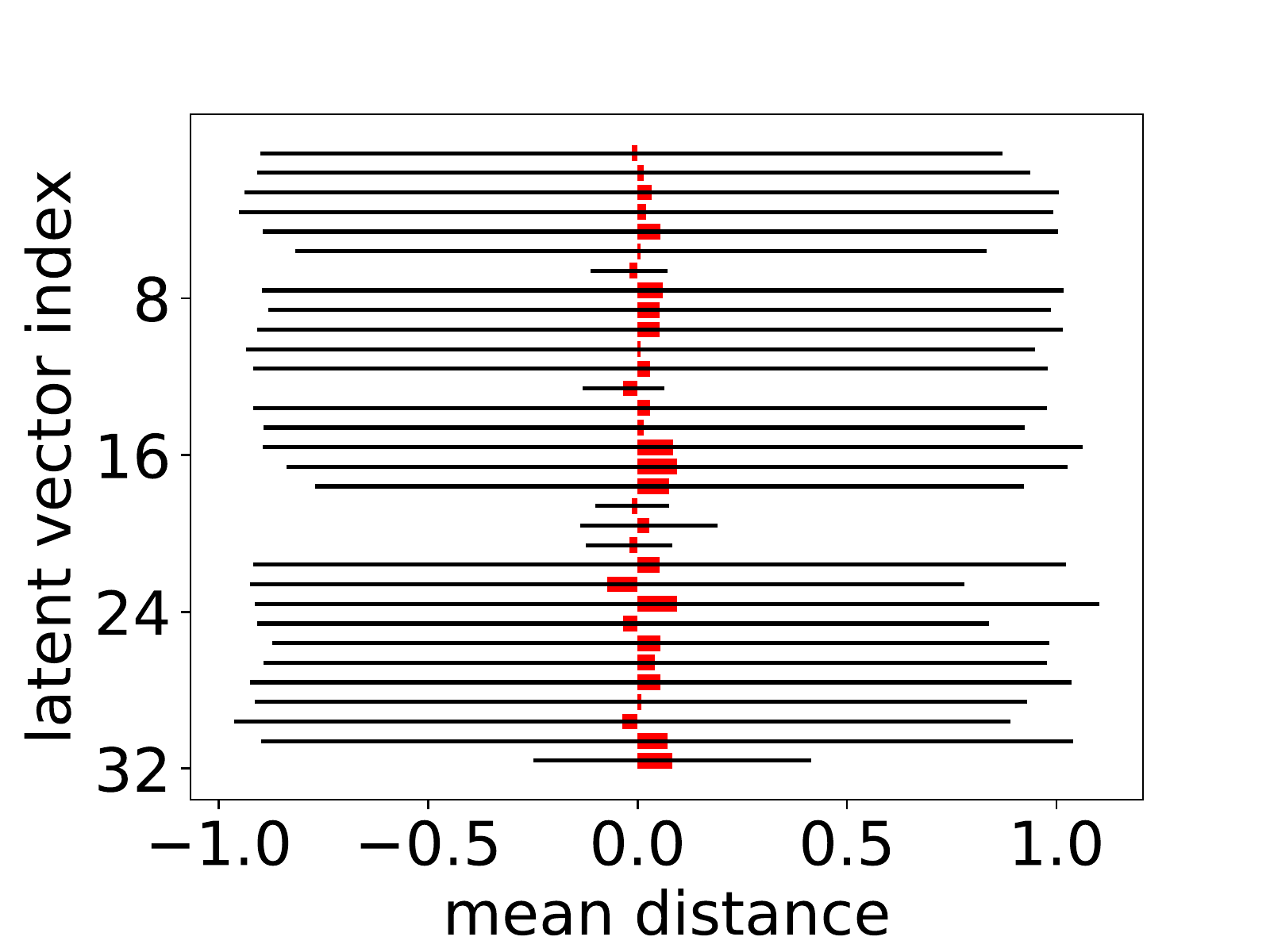}
        \caption{mean $L_1$ distance}
       \label{fig:distance_z}
    \end{subfigure}
    \caption{Difference between latent vectors $z_o$ and $z_m$ of corresponding states from vision models of $\Gamma_o$ and $\Gamma_m$.}
    \label{fig:z_diff}
\end{figure}

\subsection{Analyzing the Shared Dynamics}
To explicitly demonstrate the ability of meta-world models to unify the representation space of different environments, we provide the visualized results on evaluation set in Appendix. A.
To further investigate the shared dynamics among different environments,
we analyze the difference between latent representations $z_o$ and $z_m$ from two environments $\Gamma_o$ and $\Gamma_m$ as shown in Fig. \ref{fig:z_diff}.

Fig. \ref{fig:logstd_z} illustrates the log standard deviations of the vision models of $\Gamma_o$ and $\Gamma_m$ respectively,
where a small group of elements (indexed as 7, 13, 19, 20 and 21) has relatively low variance in both environments, thus keeping relatively stable value across different observations.
Fig. \ref{fig:distance_z} shows the mean $L_1$ distance of each element from latent representations of corresponding states between $\Gamma_o$ and $\Gamma_m$,
where the elements in the same group are close to each other in the representation space between different environments.
Being stable through different observations, these elements also share high similarity between both environments and can be interpreted mutually by both visual models.
We consider these elements as the critical part in representing the shared dynamics and define them as \emph{key elements}.

\begin{figure}[t]
    \centering
         \begin{subfigure}{0.23\textwidth}
        \includegraphics[width=0.9\textwidth]{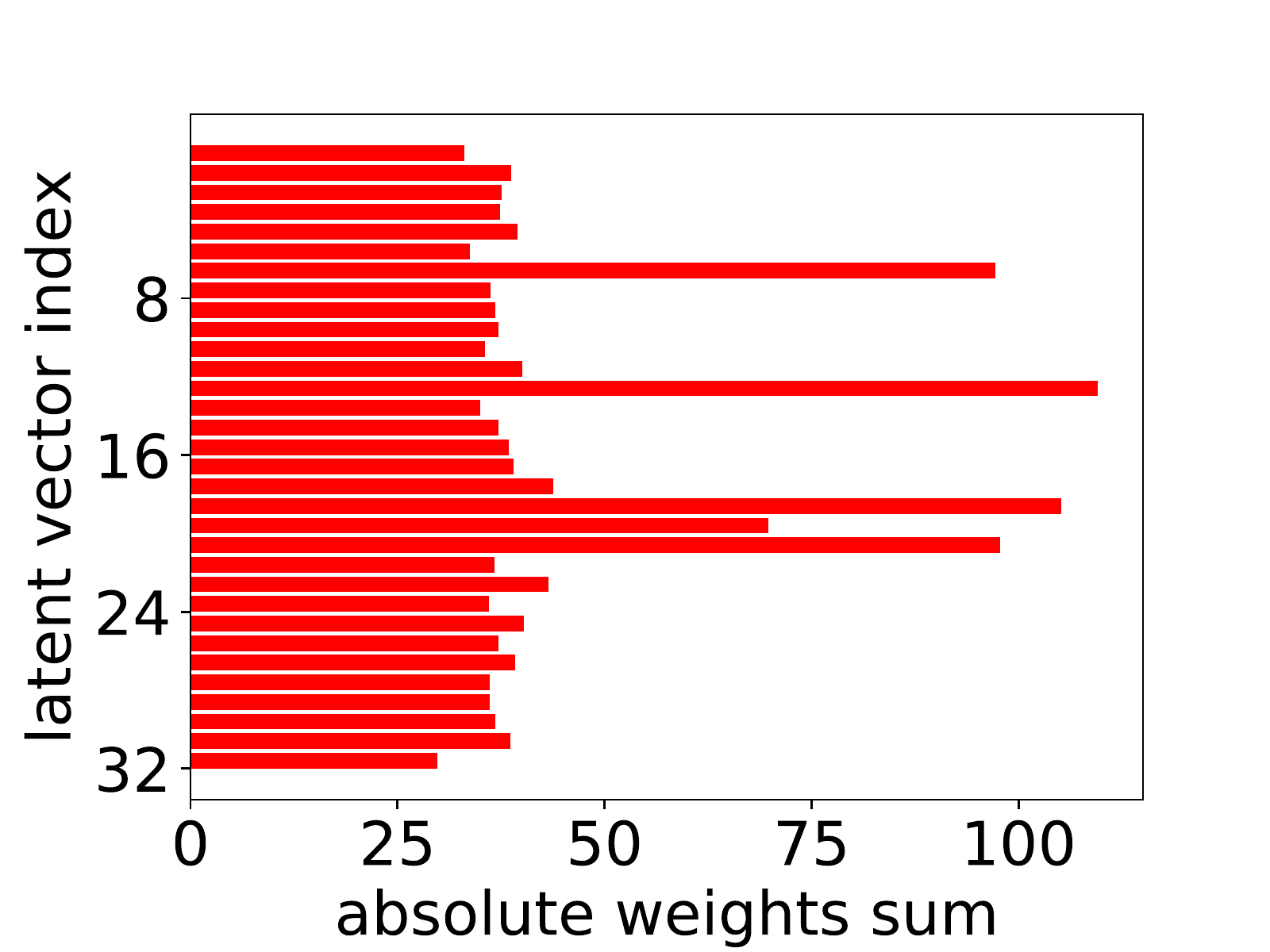}
        \caption{sum of absolute weights connected to $z_o$ in $f_o^d$}
    	\label{fig:dist_w}
    \end{subfigure}
        \label{code visualization}
     \begin{subfigure}{0.23\textwidth}
        \includegraphics[width=0.9\textwidth]{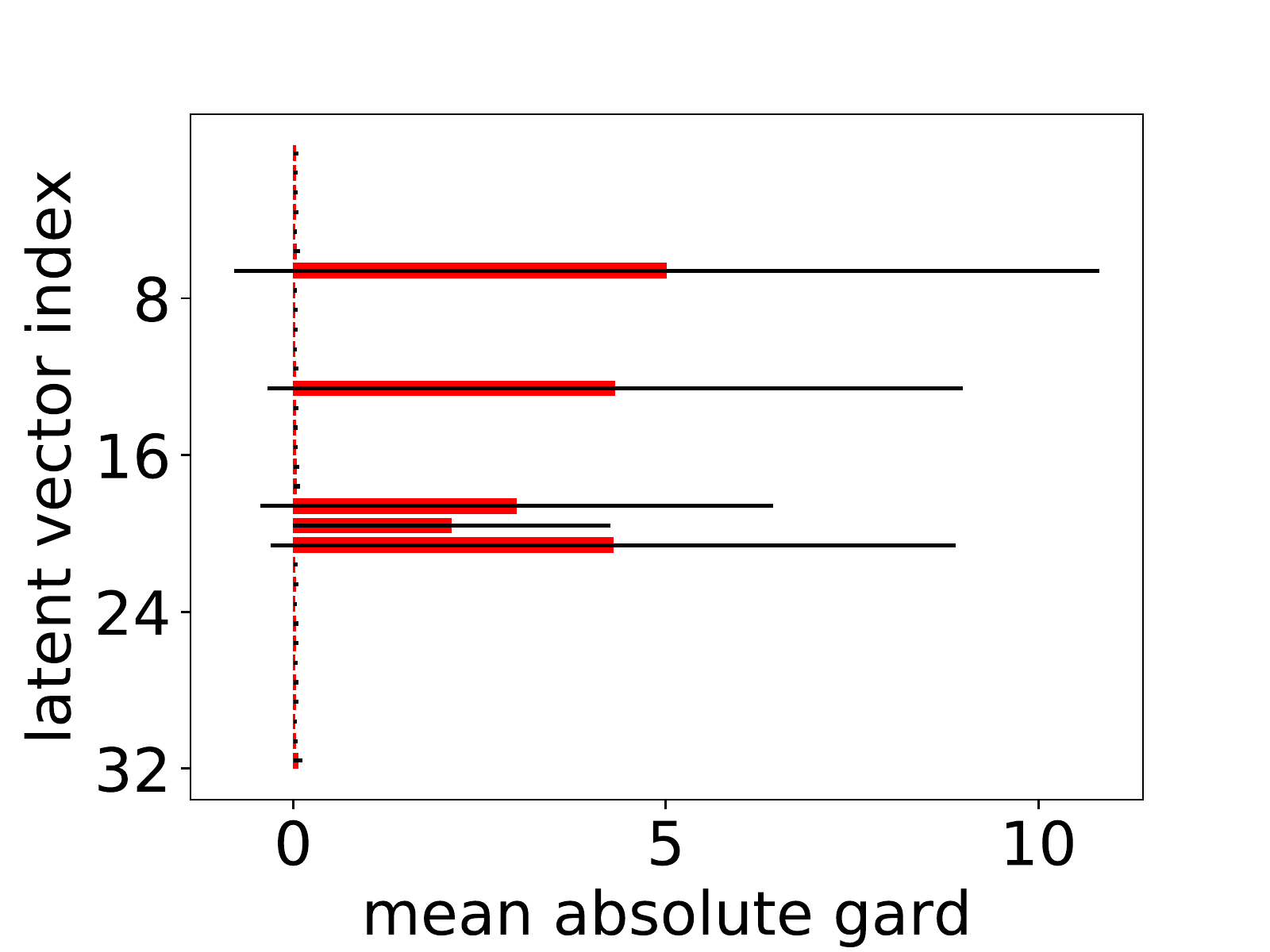}\
        \caption{mean absolute gradients of output to $z_o$ in $f_o^d$}
        \label{fig:grad_w}
    \end{subfigure}
    \caption{Validating the importance of \emph{key elements}.}
    \label{fig:key_z}
\end{figure}

To validate the importance of key elements in extracting the underlying dynamics,
we show the weights of the decoder neural network $f_o^d$ connected to the latent representations $z_o$ in Fig. \ref{fig:dist_w}.
As the sum of absolute weights connected to key elements are much larger than others,
the change of key elements will apply higher influence to the decoder output $\hat{s}_o$,
thus illustrating their significance in latent representations.
Also, the value of each element of the decoder output $\hat{s}_o$ ranges from 0 to 1,
where pixels of paddles correspond to high values.
We then compute the gradients of the output $\hat{s}_o$ with respect to the latent vector to observe which part of the latent vector contributes to the paddle more.
As shown in Fig. \ref{fig:grad_w}, the mean absolute gradients of key elements are significantly larger than others.
Indeed, other elements have nearly zero gradients,
meaning almost no contributions to the decoder output $\hat{s}_o$.
Consequently, the learned unified representations mainly concentrate on key elements.

\subsection{Self-Awareness}
One of the simple definitions of self-awareness is the ability to recognize oneself as an individual separated from the environment.
The mirror self-recognition test (MSR) is the classical method for attempting to measure self-awareness \cite{Gallup86},
where an animal is exposed to mirrors for the first time to test if it can recognize the reflected image as itself, rather than of another animal.
Humans are known to have self-consciousness as they can learn to recognize their own images after prolonged confrontation with mirrors and stop responding socially to the reflection.
Although the reason why animals can achieve this without seeing themselves before is still under debate, we propose a possible direction;
we think animals with self-consciousness should have the ability to correlate their own actions to that in the mirror and perceive the reflection as themselves.
The influence of their actions will have the same impact in both the mirrored environment and the real world even with different visual observations,
which means two environments share the same underlying dynamics.



In our experiments, we define the self-awareness of agents as the ability to unify the concepts of the agent itself in two different environments, e.g., $\Gamma_o$ and $\Gamma_m$.
Our first experiment with corresponding inputs can be seen as the Atari Pong version of the mirror test.
When taking an action in $\Gamma_o$,
an agent with meta-world models can predict how will the mirrored world $\Gamma_m$ change with the influence of this action.
Being able to predict the next observation in a mirrored Pong game,
an agent possesses the ability to distinguish between the controlled paddle and the opponent paddle by observing their actions and dynamics,
thus recognizing themselves from the reflected mirrored environment.

Meanwhile, the experiment with non-corresponding inputs takes a step further,
where agents try to figure out the correspondence between two environments without knowing the exact  transformations.
This can be seen as the process of reasoning about the differences between observations and searching the common rules among them.
Humans normally understand the world in this way,
where the process of integrating knowledge involves finding the shared aspects of different observations that progress independently.
Once the common points are found,
the conception about these observations can get unified in a certain way,
thus helping human to enhance their knowledge to better understand the world.


\section{Conclusions}
In this paper, we propose the meta-world models for learning shared underlying dynamics among visually different environments.
To propose an engineering approach to mimic the activity of human brains, which keeps a mental model of the world developed from the most impoverished of visual stimuli, we explore building a meta-world model with basic form of consciousness.
We show through extensive experiments that our meta-world model can successfully learn an abstract description of shared physical dynamics among different variants of the Atari Pong game, which can be totally different in both state space and transition functions. 
We also demonstrate that agents equipped with our meta-world model possess the ability of visual self-recognition, i.e., pass the classic mirror self-recognition test (MSR).
For future directions, we would like to explore the ability of our meta-world model understanding the world by applying it to more diverse environments.
We would also like to combine the meta-world model with "model-free" methods to learn through experience and make inferences and planning more computationally efficient.

\small
\bibliography{references}

\begin{thebibliography}{}

\bibitem[\protect\citeauthoryear{Al-Shedivat \bgroup et al\mbox.\egroup
  }{2017}]{al2017continuous}
Al-Shedivat, M.; Bansal, T.; Burda, Y.; Sutskever, I.; Mordatch, I.; and
  Abbeel, P.
\newblock 2017.
\newblock Continuous adaptation via meta-learning in nonstationary and
  competitive environments.
\newblock {\em arXiv preprint arXiv:1710.03641}.

\bibitem[\protect\citeauthoryear{Auer, Cesa-Bianchi, and
  Fischer}{2002}]{Auer2002}
Auer, P.; Cesa-Bianchi, N.; and Fischer, P.
\newblock 2002.
\newblock Finite-time analysis of the multiarmed bandit problem.
\newblock {\em Machine Learning} 47(2):235--256.

\bibitem[\protect\citeauthoryear{Baker \bgroup et al\mbox.\egroup
  }{2017}]{BakerRational17}
Baker, C.~L.; Jara-Ettinger, J.; Saxe, R.; and Tenenbaum, J.~B.
\newblock 2017.
\newblock Rational quantitative attribution of beliefs, desires and percepts in
  human mentalizing.
\newblock {\em Nature Human Behaviour} 1:0064 EP --.

\bibitem[\protect\citeauthoryear{Bellemare \bgroup et al\mbox.\egroup
  }{2015}]{Bellemare:2015:ALE:2832747.2832830}
Bellemare, M.~G.; Naddaf, Y.; Veness, J.; and Bowling, M.
\newblock 2015.
\newblock The arcade learning environment: An evaluation platform for general
  agents.
\newblock IJCAI'15,  4148--4152.
\newblock AAAI Press.

\bibitem[\protect\citeauthoryear{Clavera \bgroup et al\mbox.\egroup
  }{2018}]{clavera2018learning}
Clavera, I.; Nagabandi, A.; Fearing, R.~S.; Abbeel, P.; Levine, S.; and Finn,
  C.
\newblock 2018.
\newblock Learning to adapt: Meta-learning for model-based control.
\newblock {\em arXiv preprint arXiv:1803.11347}.

\bibitem[\protect\citeauthoryear{Dehaene, Lau, and Kouider}{2017}]{Dehaene486}
Dehaene, S.; Lau, H.; and Kouider, S.
\newblock 2017.
\newblock What is consciousness, and could machines have it?
\newblock {\em Science} 358(6362):486--492.

\bibitem[\protect\citeauthoryear{Duan \bgroup et al\mbox.\egroup
  }{2016}]{duan2016rl}
Duan, Y.; Schulman, J.; Chen, X.; Bartlett, P.~L.; Sutskever, I.; and Abbeel,
  P.
\newblock 2016.
\newblock Rl $^{}2$: Fast reinforcement learning via slow reinforcement
  learning.
\newblock {\em arXiv preprint arXiv:1611.02779}.

\bibitem[\protect\citeauthoryear{Finn, Abbeel, and
  Levine}{2017}]{finn2017model}
Finn, C.; Abbeel, P.; and Levine, S.
\newblock 2017.
\newblock Model-agnostic meta-learning for fast adaptation of deep networks.
\newblock {\em arXiv preprint arXiv:1703.03400}.

\bibitem[\protect\citeauthoryear{Forrester}{1971}]{Forrester1971}
Forrester, J.~W.
\newblock 1971.
\newblock Counterintuitive behavior of social systems.
\newblock {\em Theory and Decision} 2(2):109--140.

\bibitem[\protect\citeauthoryear{Frans \bgroup et al\mbox.\egroup
  }{2017}]{frans2017meta}
Frans, K.; Ho, J.; Chen, X.; Abbeel, P.; and Schulman, J.
\newblock 2017.
\newblock Meta learning shared hierarchies.
\newblock {\em arXiv preprint arXiv:1710.09767}.

\bibitem[\protect\citeauthoryear{Gallup}{1970}]{Gallup86}
Gallup, G.~G.
\newblock 1970.
\newblock Chimpanzees: Self-recognition.
\newblock {\em Science} 167(3914):86--87.

\bibitem[\protect\citeauthoryear{Graves \bgroup et al\mbox.\egroup
  }{2016}]{GravesHybrid2016}
Graves, A.; Wayne, G.; Reynolds, M.; Harley, T.; Danihelka, I.;
  Grabska-Barwi{\'n}ska, A.; Colmenarejo, S.~G.; Grefenstette, E.; Ramalho, T.;
  Agapiou, J.; Badia, A.; Hermann, K.~M.; Zwols, Y.; Ostrovski, G.; Cain, A.;
  King, H.; Summerfield, C.; Blunsom, P.; Kavukcuoglu, K.; and Hassabis, D.
\newblock 2016.
\newblock Hybrid computing using a neural network with dynamic external memory.
\newblock {\em Nature} 538:471 EP --.

\bibitem[\protect\citeauthoryear{Gretton \bgroup et al\mbox.\egroup
  }{2007}]{gretton2007kernel}
Gretton, A.; Borgwardt, K.~M.; Rasch, M.; Sch{\"o}lkopf, B.; and Smola, A.~J.
\newblock 2007.
\newblock A kernel method for the two-sample-problem.
\newblock In {\em Advances in NIPS},  513--520.

\bibitem[\protect\citeauthoryear{Gu \bgroup et al\mbox.\egroup
  }{2016}]{gu2016continuous}
Gu, S.; Lillicrap, T.; Sutskever, I.; and Levine, S.
\newblock 2016.
\newblock Continuous deep q-learning with model-based acceleration.
\newblock In {\em ICML},  2829--2838.

\bibitem[\protect\citeauthoryear{Ha and Schmidhuber}{2018}]{ha2018world}
Ha, D., and Schmidhuber, J.
\newblock 2018.
\newblock World models.
\newblock {\em arXiv preprint arXiv:1803.10122}.

\bibitem[\protect\citeauthoryear{Heider and Simmel}{1944}]{10.2307/1416950}
Heider, F., and Simmel, M.
\newblock 1944.
\newblock An experimental study of apparent behavior.
\newblock {\em The American Journal of Psychology} 57(2):243--259.

\bibitem[\protect\citeauthoryear{Hochreiter and
  Schmidhuber}{1997}]{Hochreiter:1997:LSM:1246443.1246450}
Hochreiter, S., and Schmidhuber, J.
\newblock 1997.
\newblock Long short-term memory.
\newblock {\em Neural Comput.} 9(8):1735--1780.

\bibitem[\protect\citeauthoryear{Kingma and Welling}{2013}]{kingma2013auto}
Kingma, D.~P., and Welling, M.
\newblock 2013.
\newblock Auto-encoding variational bayes.
\newblock {\em arXiv preprint arXiv:1312.6114}.

\bibitem[\protect\citeauthoryear{Lake \bgroup et al\mbox.\egroup
  }{2017}]{lake_ullman_tenenbaum_gershman_2017}
Lake, B.~M.; Ullman, T.~D.; Tenenbaum, J.~B.; and Gershman, S.~J.
\newblock 2017.
\newblock Building machines that learn and think like people.
\newblock {\em Behavioral and Brain Sciences} 40:e253.

\bibitem[\protect\citeauthoryear{Lake, Salakhutdinov, and
  Tenenbaum}{2015}]{Lake1332}
Lake, B.~M.; Salakhutdinov, R.; and Tenenbaum, J.~B.
\newblock 2015.
\newblock Human-level concept learning through probabilistic program induction.
\newblock {\em Science} 350(6266):1332--1338.

\bibitem[\protect\citeauthoryear{Lake}{2014}]{lake2014towards}
Lake, B.~M.
\newblock 2014.
\newblock {\em Towards more human-like concept learning in machines:
  Compositionality, causality, and learning-to-learn}.
\newblock Ph.D. Dissertation, Massachusetts Institute of Technology.

\bibitem[\protect\citeauthoryear{Leibfried, Kushman, and
  Hofmann}{2016}]{leibfried2016deep}
Leibfried, F.; Kushman, N.; and Hofmann, K.
\newblock 2016.
\newblock A deep learning approach for joint video frame and reward prediction
  in atari games.
\newblock {\em arXiv preprint arXiv:1611.07078}.

\bibitem[\protect\citeauthoryear{Levine and Abbeel}{2014}]{levine2014learning}
Levine, S., and Abbeel, P.
\newblock 2014.
\newblock Learning neural network policies with guided policy search under
  unknown dynamics.
\newblock In {\em Advances in Neural Information Processing Systems},
  1071--1079.

\bibitem[\protect\citeauthoryear{Mnih \bgroup et al\mbox.\egroup
  }{2015}]{mnih2015human}
Mnih, V.; Kavukcuoglu, K.; Silver, D.; Rusu, A.~A.; Veness, J.; Bellemare,
  M.~G.; Graves, A.; Riedmiller, M.; Fidjeland, A.~K.; Ostrovski, G.; et~al.
\newblock 2015.
\newblock Human-level control through deep reinforcement learning.
\newblock {\em Nature} 518(7540):529.

\bibitem[\protect\citeauthoryear{Parisotto, Ba, and
  Salakhutdinov}{2015}]{parisotto2015actor}
Parisotto, E.; Ba, J.~L.; and Salakhutdinov, R.
\newblock 2015.
\newblock Actor-mimic: Deep multitask and transfer reinforcement learning.
\newblock {\em arXiv preprint arXiv:1511.06342}.

\bibitem[\protect\citeauthoryear{Schmidhuber}{2015}]{schmidhuber2015learning}
Schmidhuber, J.
\newblock 2015.
\newblock On learning to think: Algorithmic information theory for novel
  combinations of reinforcement learning controllers and recurrent neural world
  models.
\newblock {\em arXiv preprint arXiv:1511.09249}.

\bibitem[\protect\citeauthoryear{Spelke}{1990}]{SPELKE199029}
Spelke, E.~S.
\newblock 1990.
\newblock Principles of object perception.
\newblock {\em Cognitive Science} 14(1):29 -- 56.

\bibitem[\protect\citeauthoryear{Sutton}{1990}]{sutton1990integrated}
Sutton, R.~S.
\newblock 1990.
\newblock Integrated architectures for learning, planning, and reacting based
  on approximating dynamic programming.
\newblock In {\em Machine Learning Proceedings 1990}. Elsevier.
\newblock  216--224.

\bibitem[\protect\citeauthoryear{Teh \bgroup et al\mbox.\egroup
  }{2017}]{teh2017distral}
Teh, Y.; Bapst, V.; Czarnecki, W.~M.; Quan, J.; Kirkpatrick, J.; Hadsell, R.;
  Heess, N.; and Pascanu, R.
\newblock 2017.
\newblock Distral: Robust multitask reinforcement learning.
\newblock In {\em Advances in NIPS},  4496--4506.

\bibitem[\protect\citeauthoryear{Turing}{1950}]{doi:10.1093/mind/LIX.236.433}
Turing, A.~M.
\newblock 1950.
\newblock Computing machinery and intelligence.
\newblock {\em Mind} LIX(236):433--460.

\bibitem[\protect\citeauthoryear{Van~Gulick}{2018}]{sep-consciousness}
Van~Gulick, R.
\newblock 2018.
\newblock Consciousness.
\newblock In Zalta, E.~N., ed., {\em The Stanford Encyclopedia of Philosophy}.
  Metaphysics Research Lab, Stanford University, spring 2018 edition.

\bibitem[\protect\citeauthoryear{Wahlstr{\"o}m, Sch{\"o}n, and
  Deisenroth}{2015}]{wahlstrom2015pixels}
Wahlstr{\"o}m, N.; Sch{\"o}n, T.~B.; and Deisenroth, M.~P.
\newblock 2015.
\newblock From pixels to torques: Policy learning with deep dynamical models.
\newblock {\em arXiv preprint arXiv:1502.02251}.

\bibitem[\protect\citeauthoryear{Wang \bgroup et al\mbox.\egroup
  }{2016}]{wang2016learning}
Wang, J.~X.; Kurth-Nelson, Z.; Tirumala, D.; Soyer, H.; Leibo, J.~Z.; Munos,
  R.; Blundell, C.; Kumaran, D.; and Botvinick, M.
\newblock 2016.
\newblock Learning to reinforcement learn.
\newblock {\em arXiv preprint arXiv:1611.05763}.

\bibitem[\protect\citeauthoryear{Watter \bgroup et al\mbox.\egroup
  }{2015}]{watter2015embed}
Watter, M.; Springenberg, J.; Boedecker, J.; and Riedmiller, M.
\newblock 2015.
\newblock Embed to control: A locally linear latent dynamics model for control
  from raw images.
\newblock In {\em Advances in NIPS},  2746--2754.

\bibitem[\protect\citeauthoryear{Wellman and
  Gelman}{1992}]{doi:10.1146/annurev.ps.43.020192.002005}
Wellman, H.~M., and Gelman, S.~A.
\newblock 1992.
\newblock Cognitive development: Foundational theories of core domains.
\newblock {\em Annual Review of Psychology} 43(1):337--375.
\newblock PMID: 1539946.

\end{thebibliography}
\bibliographystyle{aaai}
\onecolumn
\appendix
\section{A. Unifying the representation space of different environments}
\begin{figure*}[htbp]
    \centering    
     \begin{subfigure}{0.11\textwidth}
        \makebox[\textwidth]{\raisebox{0.88\textwidth}{$t=0$}}\\
        \makebox[\textwidth]{\raisebox{0.88\textwidth}{$t=1$}}\\
        \makebox[\textwidth]{\raisebox{0.88\textwidth}{$t=2$}}\\
        \makebox[\textwidth]{\raisebox{0.88\textwidth}{$t=3$}}\\
        \makebox[\textwidth]{\raisebox{0.88\textwidth}{$t=4$}}\\
        \makebox[\textwidth]{\raisebox{0.88\textwidth}{$t=5$}}\\
        \makebox[\textwidth]{\raisebox{0.88\textwidth}{$t=6$}}\\
        \makebox[\textwidth]{\raisebox{0.88\textwidth}{$t=7$}}\\
    \end{subfigure}
     \begin{subfigure}{0.11\textwidth}
        \includegraphics[width=\textwidth]{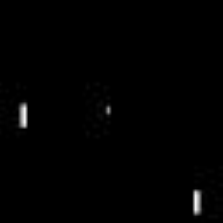}
        \includegraphics[width=\textwidth]{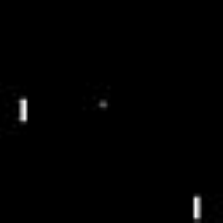}
        \includegraphics[width=\textwidth]{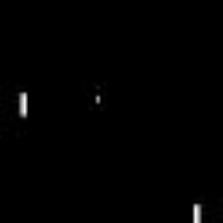}
        \includegraphics[width=\textwidth]{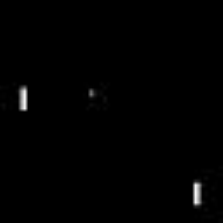}
        \includegraphics[width=\textwidth]{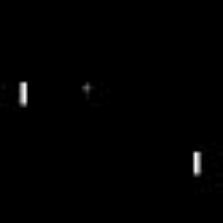}
        \includegraphics[width=\textwidth]{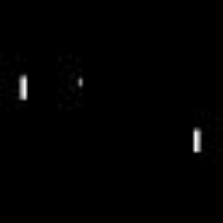}
        \includegraphics[width=\textwidth]{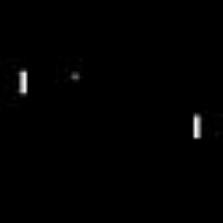}
        \includegraphics[width=\textwidth]{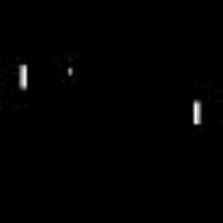}
        \caption{$\Gamma_o$}
    \end{subfigure}
    \begin{subfigure}{0.11\textwidth}
        \includegraphics[width=\textwidth]{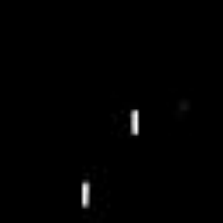}
        \includegraphics[width=\textwidth]{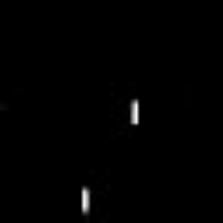}
        \includegraphics[width=\textwidth]{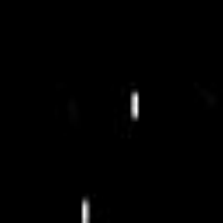}
        \includegraphics[width=\textwidth]{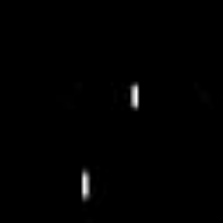}
        \includegraphics[width=\textwidth]{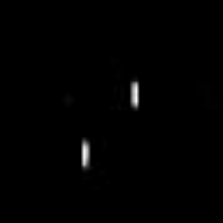}
        \includegraphics[width=\textwidth]{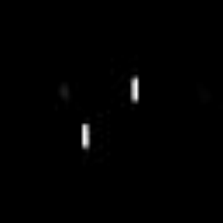}
        \includegraphics[width=\textwidth]{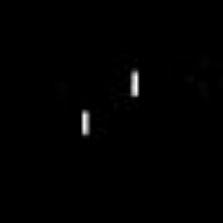}
        \includegraphics[width=\textwidth]{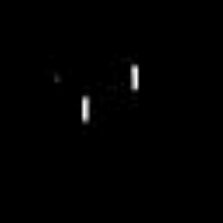}
        \caption{$\Gamma_h$}
    \end{subfigure}
    \begin{subfigure}{0.11\textwidth}
        \includegraphics[width=\textwidth]{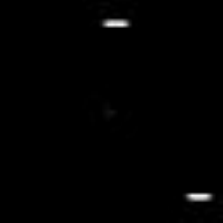}
        \includegraphics[width=\textwidth]{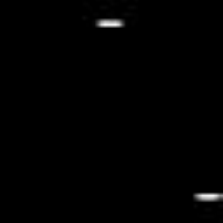}
        \includegraphics[width=\textwidth]{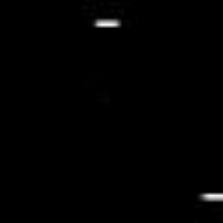}
        \includegraphics[width=\textwidth]{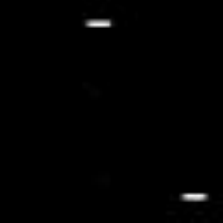}
        \includegraphics[width=\textwidth]{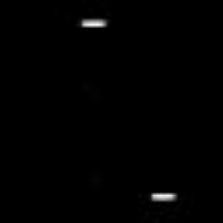}
        \includegraphics[width=\textwidth]{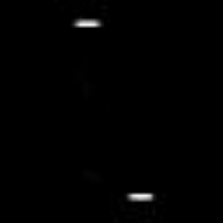}
        \includegraphics[width=\textwidth]{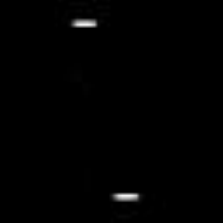}
        \includegraphics[width=\textwidth]{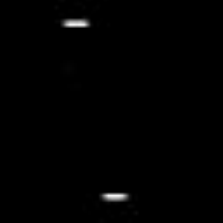}
        \caption{$\Gamma_t$}
    \end{subfigure}
    \begin{subfigure}{0.11\textwidth}
        \includegraphics[width=\textwidth]{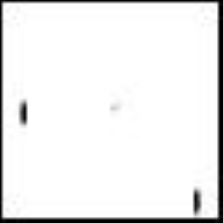}
        \includegraphics[width=\textwidth]{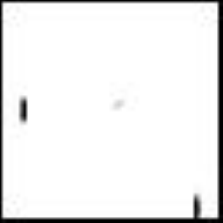}
        \includegraphics[width=\textwidth]{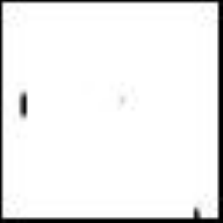}
        \includegraphics[width=\textwidth]{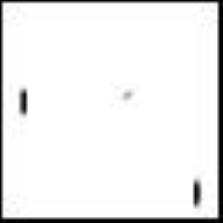}
        \includegraphics[width=\textwidth]{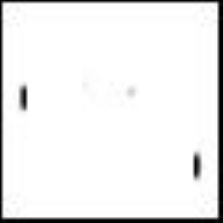}
        \includegraphics[width=\textwidth]{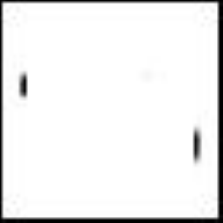}
        \includegraphics[width=\textwidth]{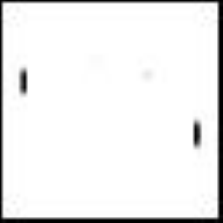}
        \includegraphics[width=\textwidth]{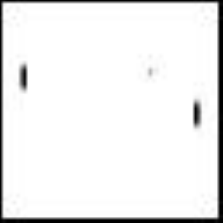}
        \caption{$\Gamma_c$}
    \end{subfigure}
    \begin{subfigure}{0.11\textwidth}
        \includegraphics[width=\textwidth]{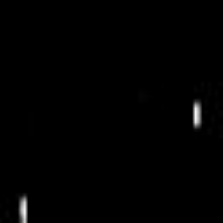}
        \includegraphics[width=\textwidth]{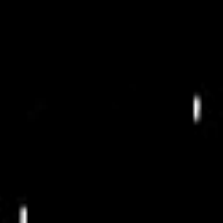}
        \includegraphics[width=\textwidth]{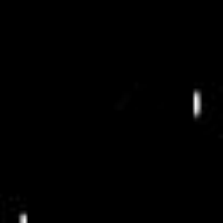}
        \includegraphics[width=\textwidth]{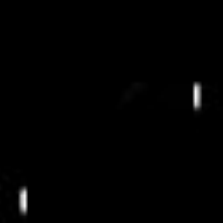}
        \includegraphics[width=\textwidth]{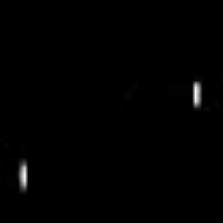}
        \includegraphics[width=\textwidth]{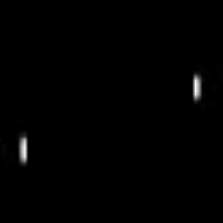}
        \includegraphics[width=\textwidth]{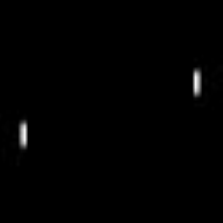}
        \includegraphics[width=\textwidth]{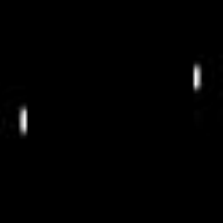}
        \caption{$\Gamma_m$}
    \end{subfigure}
    \begin{subfigure}{0.11\textwidth}
        \includegraphics[width=\textwidth]{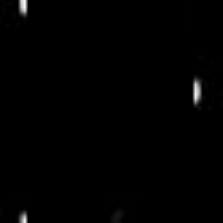}
        \includegraphics[width=\textwidth]{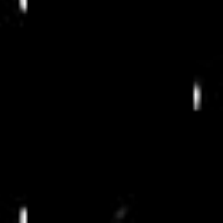}
        \includegraphics[width=\textwidth]{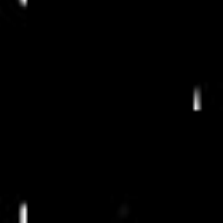}
        \includegraphics[width=\textwidth]{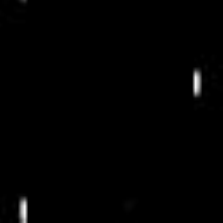}
        \includegraphics[width=\textwidth]{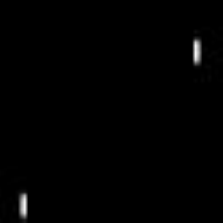}
        \includegraphics[width=\textwidth]{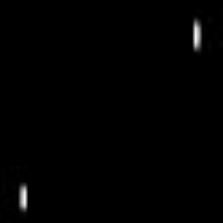}
        \includegraphics[width=\textwidth]{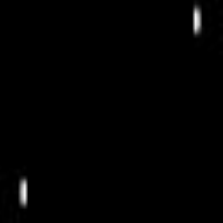}
        \includegraphics[width=\textwidth]{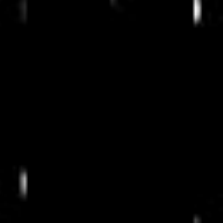}
        \caption{$\Gamma_v$}
    \end{subfigure}
    \caption{The visualized results on evaluation set. The first column is the observation of environment $\Gamma_o$, which is encoded to $z_o$ by V model $V_o$. The rest columns correspond to the decoding results of $z_o$ from vision models of $\Gamma_h$, $\Gamma_t$, $\Gamma_c$, $\Gamma_m$, $\Gamma_v$. }
    \label{fig:meta_visualization}
\end{figure*}

\end{document}